\documentclass{clv3}

\usepackage{hyperref}
\usepackage{xcolor}
\usepackage{times}
\usepackage{latexsym}
\usepackage{multirow}
\usepackage{url}
\usepackage{subfig}
\usepackage{graphicx}
\usepackage{framed}
\usepackage{amsmath}
\usepackage{amssymb}
\usepackage{amsthm}
\usepackage{mathtools}
\usepackage{float}
\definecolor{darkblue}{rgb}{0, 0, 0.5}
\hypersetup{colorlinks=true,citecolor=darkblue, linkcolor=darkblue, urlcolor=darkblue}

\issue{1}{1}{2018}

\runningtitle{Sarcasm Analysis using Conversation Context}

\runningauthor{Ghosh, Fabbri, and Muresan}

\begin{document}

\title{Sarcasm Analysis using Conversation Context}

\author{Debanjan Ghosh\thanks{The research was carried out while Debanjan was a Ph.D. candidate at Rutgers University.}}
\affil{McGovern Institute for Brain Research \\ Massachusetts Institute of Technology\\ dg513@mit.edu}

\author{Alexander R. Fabbri}
\affil{Department of Computer Science \\
Yale University \\
alexander.fabbri@yale.edu}

\author{Smaranda Muresan}
\affil{Data Science Institute \\
Columbia University \\
smara@columbia.edu}

\maketitle

\begin{abstract}
Computational models for sarcasm detection have often relied on the content of utterances in isolation. However, the speaker's sarcastic intent is not always apparent without additional context. Focusing on social media discussions, we investigate three issues: (1) does modeling conversation context help in sarcasm detection; (2) can we identify what part of conversation context triggered the sarcastic reply; and (3) given a sarcastic post that contains multiple sentences, can we identify the specific sentence that is sarcastic. To address the first issue, we investigate several types of Long Short-Term Memory (LSTM) networks that can model both the conversation context and the current turn. We show that LSTM networks with sentence-level attention on context and current turn, as well as the conditional LSTM network \cite{rocktaschel2015reasoning}, outperform the LSTM model that reads only the current turn. As conversation context, we consider the prior turn, the succeeding turn or both. Our computational models are tested on two types of social media platforms: Twitter and discussion forums. We discuss several differences between these datasets ranging from their size to the nature of the gold-label annotations. To address the last two issues, we present a qualitative analysis of the attention weights produced by the LSTM models (with attention) and discuss the results compared with human performance on the two tasks.  

\end{abstract}

\section{Introduction} \label{intro}



Social media has stimulated the production of user-generated content that contains figurative language use such as sarcasm and irony. Recognizing sarcasm and verbal irony is critical for understanding people's actual sentiments and beliefs \cite{maynard2014cares}. For instance, the utterance ``I love waiting at the doctor's office for hours \dots" is ironic, expressing a negative sentiment towards the situation of ``waiting for hours at the doctor's office", even if the speaker uses positive sentiment words such as ``love". 




Verbal irony and sarcasm are a type of interactional phenomenon with specific perlocutionary effects on the hearer \citep{haverkate1990speech}, such as to break their pattern of expectation. 
For the current paper, we do not make a clear distinction between sarcasm and verbal irony. 
Most computational models for sarcasm detection have considered utterances in isolation  \citep{davidov2010,gonzalez,liebrecht2013perfect,riloff,maynard2014cares,joshi2015,
ghoshguomuresan2015EMNLP,joshi2016word,ghosh2016fracking}. In many instances, however, even humans have difficulty in recognizing sarcastic intent when considering an utterance in isolation \cite{wallace2014humans}. Thus, to detect the speaker's sarcastic intent, it is necessary (even if maybe not sufficient) to consider their utterance(s)  in the larger \emph{conversation context}. Consider the Twitter conversation example in Table \ref{table:priorcurrent}. Without the context of userA's
statement, the sarcastic intent of userB's response might not be detected.

\begin{table}
\centering
\begin{tabular}{ p{2cm}|p{1.2cm}|p{8cm} }
\hline
Platform & Turn Type & \multicolumn{1}{c}{Turn pairs} \\
\hline
\multirow{2}{*}{Twitter} & {\sc p\_turn}  & {\bf userA:} Plane window shades are open during take-off \& landing so that people can see if there is fire in case of an accident URL . \\ & {\sc c\_turn} & {\bf userB}: @UserA \dots awesome \dots one more reason to feel really great about flying \dots \#sarcasm. \\ \hline

\multirow{2}{4em}{Discussion Forum ($IAC_{v2}$)} & {\sc p\_turn} & {\bf userC:}  how do we rationally explain these creatures existence so recently in our human history if they were extinct for millions of years? and if they were the imaginings of bronze age sheep herders as your atheists/evolutionists would have you believe, then how did these ignorant people describe creatures we can now recognize from fossil evidence? and while your at it, ask yourself if it's reasonable that the bones of dead creatures have survived from 60 million years to some estimated to be more than 200 million years without becoming dust?   \\
& {\sc c\_turn} & {\bf userD:} How about this explanation - you're reading WAAAAAY too much into your precious Bible.\\
\hline
\multirow{2}{4em}{Discussion Forum ($Reddit$)} & {\sc p\_turn} & {\bf userE:}  nothing will happen, this is going to die a quiet death like 99.99 \% of other private member motions. this whole thing is being made into a big ordeal by those that either don't know how our parliament works, or are trying to push an agenda. feel free to let your mp know how you feel though but i doubt this motion gets more than a few minutes of discussion before it is send to the trashcan. \\
& {\sc c\_turn} & {\bf userF:} the usual ``nothing to see here'' response. whew! we can sleep at night and ignore this.  \\
\hline
\end{tabular}
\caption{Sarcastic turns ({\sc c\_turn}) and their respective prior turns ({\sc p\_turn}) in Twitter and discussion forums (Internet Argument Corpus (IAC) \cite{orabycreating} and $Reddit$}
\label{table:priorcurrent}
\end{table}

\begin{table}
\centering
\begin{tabular}{ p{1.2cm}|p{10cm} }
\hline
 Turn Type & \multicolumn{1}{c }{Social media discussion} \\
\hline
{\sc p\_turn} & {\bf userA:}  my State is going to heII in a handbasket since these lefties took over. 
emoticonXBanghead.  \\
{\sc c\_turn} & {\bf userB:} Well since Bush took office the mantra has bee "Local Control" has it not. 
Apparently the people of your state want whats happening. 
Local control in action. Rejoice in your victory.\\
{\sc s\_turn} & {\bf userC:} I think the trip was a constructive idea, especially for high risk middle school youths \dots. Perhaps the program didn't respect their high risk homes enough. If it were a different group of students, the parents would have been told. The program was the YMCA, not lefty, but Christian based. \\
\hline
{\sc p\_turn} & {\bf userA:}  In his early life, X had a reputation for drinking too much. Whether or not this affected his thinking is a question which should to be considered when asking questions about mormon theology \dots
emoticonXBanghead.  \\
{\sc c\_turn} & {\bf userB:} Wow, that must be some good stuff he was drinking to keep him 'under the influence' for THAT long!! :p\\
{\sc s\_turn} & {\bf userC:} Perhaps he was stoned on other drugs like the early writers of the bible.\\
\hline
\end{tabular}
\caption{Sarcastic messages ({\sc c\_turn}s) and their respective prior turns ({\sc p\_turn}) and succeeding turns ({\sc s\_turn}) from $IAC_{v2}$.}
\label{table:priorcurrentnext}
\end{table}

In this paper, we investigate the role of \textit{conversation context} for the detection of sarcasm in social media discussions (Twitter conversations and discussion forums). The unit of analysis (i.e., what we label as sarcastic or not sarcastic) is a message/turn in a social media conversation (i.e., a tweet in Twitter or a post/comment in discussion forums). We call this unit  \emph{current turn} ({\sc c\_turn}). The conversation context that we consider is the \emph{prior turn} ({\sc p\_turn}), and, when available, also the \emph{succeeding turn} ({\sc s\_turn}), which is the reply to the \emph{current turn}. Table \ref{table:priorcurrent} shows some examples of sarcastic messages  ({\sc c\_turn}s) together with their respective prior turns ({\sc p\_turn}) taken from Twitter and two discussion forum corpora: the Internet Argument Corpus ($IAC_{v2}$) \cite{orabycreating} and Reddit \cite{khodak2017large}. 
  Table \ref{table:priorcurrentnext} shows examples from the $IAC_{v2}$ corpus of sarcastic messages ({\sc c\_turn}s; userB's post) and the conversation context given by the prior turn ({\sc p\_turn}; userA's post) as well as the succeeding turn ({\sc s\_turn}; userC's post).

We address three specific questions:
\begin{enumerate}
\item Does modeling of conversation context help in sarcasm detection?
\item Can humans and computational models identify what part of the prior turn ({\sc p\_turn}) triggered the sarcastic reply ({\sc c\_turn}) (e.g., which sentence(s) from userC's turn triggered userD's sarcastic reply in Table \ref{table:priorcurrent})?
\item Given a sarcastic message ({\sc c\_turn}) that contains multiple sentences, can humans and computational models identify the specific sentence that is sarcastic?
\end{enumerate}

To answer the first question, we consider two types of context: 1) just the prior turn and 2) both the prior and the succeeding turns. 
We investigate both Support Vector Machine models \citep{CortesVapnik1995,libsvm} with linguistically-motivated discrete features and several types of Long Short-Term Memory (LSTM) networks 
\cite{hochreiter1997long} that can model both the conversation context (i.e., {\sc p\_turn}, {\sc s\_turn} or both) and the current turn ({\sc c\_turn}) (Section \ref{section:experiment}). We utilize different flavors of the LSTM networks and we show that the conditional LSTM network \cite{rocktaschel2015reasoning} and the LSTM networks with sentence-level attention on current turn ({\sc c\_turn}) \emph{and} context (particularly the prior turn)  outperform the LSTM model that reads only the current turn ({\sc c\_turn}) (Section \ref{section:results}). We perform a detailed error analysis. 
Our computational models are tested on two different types of social media platforms: micro-blogging platform such as Twitter and discussion forums. Our datasets, introduced in Section \ref{section:corpus}, differ on two main dimensions. First,  discussion forum posts are much longer than Twitter messages, which makes them particularly relevant for the last two questions we try to address. Second, the gold labels for the sarcastic class are obtained differently: while Twitter and Reddit corpora are self-labeled (i.e., speakers themselves label their messages as sarcastic), the $IAC_{v2}$ corpus is labeled via crowdsourcing. Thus, for the latter, the gold labels emphasize whether the sarcastic intent of the speaker has been \emph{perceived} by the hearers/annotators (we do not know if the speaker intended to be sarcastic or not). We perform a study of training on Reddit data (self-labeled) and testing on $IAC_{v2}$ (labeled via crowdsourcing). 
To answer the second and third questions, we present a qualitative analysis of the attention weights produced by the LSTM models with attention and discuss the results compared with human performance on the tasks (Section \ref{section:qualitative}). We make all datasets and code available.\footnote{We use Theano Python library for the LSTM-based experiments. Code available at https://github.com/debanjanghosh/sarcasm\_context and https://github.com/Alex-Fabbri/deep\_learning\_nlp\_sarcasm/.}

\section{Related Work} \label{section:related}

Existing work in computational models for sarcasm detection addresses a variety of different tasks. These include, primarily, classifying sarcastic vs. non-sarcastic utterances using various lexical and pragmatic features \citep{gonzalez,liebrecht2013perfect,muresanjasist2016,joshi2016word,ghosh2016fracking}, rules and text-patterns \citep{veale2010detecting}, specific hashtags \citep{maynard2014cares} as well as semi-supervised approach \citep{davidov2010}. Researchers have also examined different characteristics of sarcasm, such as sarcasm detection as a sense-disambiguation problem \citep{ghoshguomuresan2015EMNLP} and sarcasm as a contrast between a positive sentiment and negative situation \citep{riloff,joshi2015}. Apart from linguistically motivated contextual knowledge, cognitive features, such as eye-tracking information, are also used in sarcasm detection \citep{mishra2016harnessing}. \citet{schifanella2016detecting} propose a multi-modal approach, where textual and visual features are combined for sarcasm detection. Some studies present approaches for sarcasm detection in languages other than English. For example, \citet{ptavcek2014sarcasm} use various n-grams, including unigrams, bigrams, trigrams and a set of language-independent features, such as punctuation marks, emoticons, quotes, capitalized words, character n-grams features to identify sarcasm in Czech tweets. Similarly, \citet{liu2014sarcasm} introduce POS sequences, homophony features to detect sarcasm from Chinese utterances. \citet{bharti2017harnessing} compared tweets written in Hindi to news context for irony identification.

Most of the above mentioned approaches have considered utterances in isolation. However, even humans have difficulty sometimes in recognizing sarcastic intent when considering an utterance in isolation \citep{wallace2014humans}. Recently an increasing number of researchers have started using contextual information for irony and sarcasm detection. The term context loosely refers to any \textit{information} that is available beyond the utterance itself \citep{joshi2016automatic}. There are two major research directions - \textit{author} context  and \textit{conversation} context, and we briefly discuss them here. 

\paragraph{Author Context} Researchers often examined the author-specific context \citep{khattri2015your,rajadesingan2015sarcasm}. For instance, \citet{khattri2015your} studied the historical tweets of a particular author to learn about the author's prevailing sentiment towards particular targets (e.g., named entities). Here, historical tweets are considered as the \textit{author's context}. \citet{khattri2015your} hypothesized that altering sentiment towards a particular target in the candidate tweet may represent sarcasm. 
\citet{rajadesingan2015sarcasm} create features based on authors' previous tweets, for instance, author's familiarity with sarcasm. Finally, \citet{amir2016modelling} enhanced \citet{rajadesingan2015sarcasm}'s model by creating user embeddings based on the tweets of users and combined that with regular utterance-based embeddings for sarcasm detection. 

\paragraph{Conversation Context} \citet{wallace2014humans} present an annotation study where the annotators identify sarcastic comments from Reddit threads and were allowed to utilize additional context for sarcasm labeling.   They also use a lexical classifier to automatically identify sarcastic comments and show that the model often fails to recognize the same examples for which the annotators requested more context. \citet{bamman2015contextualized} considered conversation context in addition to ``author and addressee'' features that are derived from the author's historical tweets, profile information and historical communication between the author and the addressee. Their results show only a minimal impact of modeling conversation context. \citet{oraby2017you} have studied the ``pre'' and ``post'' messages from debate forums as well as Twitter to identify whether Rhetorical Question are used sarcastically or not. For both corpora adding `pre'' and ``post'' messages do not seem to affect significantly the F1 scores, even though using the ``post" message as context seems to improve for the sarcastic class \citep{oraby2017you}. Unlike the above approaches that model the utterance and context together, \citet{wang2015twitter} and \citet{joshi2016harnessing} use a sequence labeling approach and show that conversation helps in sarcasm detection.  Inspired by this idea of modeling the current turn and context separately, in our prior work \citep{ghoshsigdial2017} --- which this paper substantially extends ---, we proposed a deep learning architecture based on LSTMs, where one LSTM reads the context (prior turn) and one LSTM reads the current turn, and showed that this type of architecture outperforms a simple LSTM that just reads the current turn. Independently, \citet{ghosh2017magnets} have proposed a similar architecture based on Bi-LSTMs to detect sarcasm in Twitter. Unlike \citet{ghosh2017magnets}, our prior work used attention-based LSTMs that allowed us to investigate whether we can identify what part of the conversation context triggered the sarcastic reply, and showed results both on discussion forum data and Twitter.    

This paper substantially extends our prior work introduced in \citet{ghoshsigdial2017}.  
First, we extend the notion of context to consider also the ``succeeding turn'' not only the ``prior turn'' and for that we collected a subcorpus from the Internet Argument Corpus (IAC) that contains both the prior turn and the succeeding turn as context. Second, we present a discussion on the nature of the datasets in terms of size and how the gold labels are obtained (self-labeled vs. crowdsource labeled), which might provide insights into the nature of sarcasm in social media. We use a new discussion forum dataset from Reddit that is another example of a self-labeled dataset (besides Twitter), where the speakers label their own post as sarcastic using the ``/s" marker. We present an experiment where we train on the Reddit dataset (self-labeled data) and test on IAC (where the gold labels were assigned via crowdsourcing). Third, we present a detailed error analysis of the computational models. Fourth, we address a new question: given a sarcastic message that contains multiple sentences, can humans and computational models identify the specific sentence that is sarcastic? We conduct comparative analysis  between human performance on the task and the attention weights of the LSTM models. In addition, for all the crowdsourcing experiments we include more details on the inter-annotators agreement among Turkers. Fifth, we include new baselines (tf-idf; RBF kernels) and a run using unbalanced datasets. Last but not least, we empirically show that explicitly modeling the turns helps and provides better results than just concatenating the current turn and prior turn (and/or succeeding turn). This experimental result supports the conceptual claim that both we and \citet{ghosh2017magnets} make that it is important to keep the C\_TURN and the P\_TURN (S\_TURN) separate (e.g., modeled by different LSTMs), as the model is designed to recognize a possible inherent incongruity between them. This incongruity might become diffuse if the inputs are combined too soon (i.e., using one LSTM on combined current turn and context). 


\paragraph{LSTM for Natural Language Inference (NLI) tasks and sarcasm detection}
 Long Short-Term Memory (LSTM) networks are a particular type of recurrent neural networks that have been shown to be effective in NLI tasks, especially where the task is to establish the relationship between multiple inputs. For instance, in recognizing textual entailment research, LSTM networks, especially the attention-based models, are highly accurate \citep{rocktaschel2015reasoning,bowman2015large,parikh2016decomposable,sha2016reading}. \citet{rocktaschel2015reasoning} presented various word-based and conditional attention models that show how the entailment relationship between the \textit{hypothesis} and the \textit{premise} can be effectively derived. \citet{parikh2016decomposable} uses attention to decompose the RTE problem into subproblems that can be solved separately and \citep{sha2016reading} presented an altered version (``re-read LSTM'') of LSTM that is similar to word attention models of \citet{rocktaschel2015reasoning}. Likewise, recently LSTMs are used in sarcasm detection research \citep{oraby2017you,huang2017irony,ghosh2017magnets}. \citet{oraby2017you} used LSTM models to identify sarcastic utterances (tweets and posts from the $IAC_{v2}$ that are structured as rhetorical questions),  \citet{huang2017irony} applied LSTM for sense-disambiguation research (on the same dataset proposed by \citet{ghoshguomuresan2015EMNLP}), and \citet{ghosh2017magnets} used bi-directional LSTMs to identify sarcastic tweets. In our research, we use multiple LSTMs for each text units, e.g.,  the context and the response. We observe that the LSTM$^{conditional}$ model and the sentence level attention-based models using both context and reply present the best results.

\section{Data} \label{section:corpus}

One goal of our investigation is to comparatively study two types of social media platforms that have been considered individually for sarcasm detection:  discussion forums and Twitter. In addition, the choice of our datasets reflects another critical aspect: the nature of the gold label annotation of sarcastic messages. On the one hand, we have self-labeled data (i.e., the speakers themselves labeled their posts as sarcastic) in the case of Twitter and Reddit data. On the other hand, we have labels obtained via crowdsourcing as is the case for the Internet Argument Corpus \cite{orabycreating}. We first introduce the different datasets we use and then point out some differences between them that could impact results and modeling choices.

\paragraph{{\bf Internet Argument Corpus V2 ($IAC_{v2}$)}} Internet Argument Corpus ($IAC$) is a publicly available corpus of online forum conversations on a range of social and political topics, from gun control debates and marijuana legalization to climate change and evolution \citep{walker2012corpus}. The corpus comes with annotations of different types of pragmatic categories such as agreement/disagreement (between a pair of online posts), nastiness, and sarcasm. There are different version of $IAC$ and we use a specific subset of $IAC$ in this research. \citet{orabycreating} have introduced a subset of the Internet Argument Corpus V2 that contains 9,400 posts labeled as sarcastic or non-sarcastic, called Sarcasm Corpus V2 (balanced dataset). To obtain the gold labels, \citet{orabycreating} first employed a weakly supervised pattern learner to learn sarcastic and non-sarcastic patterns from the $IAC$ posts and later 
employ a multiple stage crowdsourcing process to identify sarcastic and non-sarcastic posts. Annotators were asked to label a post (current turn ({\sc c\_turn}) in our terminology) as sarcastic if any part of the post  contained sarcasm, and thus the annotation is done at the comment level and not the sentence level. This dataset contains the post ({\sc c\_turn}) as well as the quotes to which the posts are replies to (i.e., prior turn ({\sc p\_turn}) in our terminology). Sarcasm annotation was based on identifying three types of sarcasm; (a) general (i.e., mostly based on lexico-syntactic patterns) (b) rhetorical questions (i.e., questions that are not information seeking questions but formed as an indirect assertion \citep{frank1990you}; denoted as $RQ$) and (c) use of hyperbolic terms (i.e., use of ``best'',``greatest'', ``nicest'', etc. \citep{camp2011}). Although the dataset described by \citet{orabycreating} consists of 9,400 posts, only 50\%  of that corpus is currently available for research (4,692 altogether; balanced between sarcastic and non-sarcastic categories while maintaining the same distribution of general, hyperbolic, or $RQ$ type sarcasm). This is the dataset we used in our study and denote as $IAC_{v2}$.\footnote{\citet{orabycreating} reported best F1 scores between 65\% to 74\% for the three types of sarcasm. However, the reduction in the training size of the released corpus might negatively impact the classification performance.} Table \ref{table:priorcurrent} shows an example of sarcastic current turn (userD's post) and its prior turn (userC's post) from the $IAC_{v2}$ dataset. 

The $IAC_{v2}$ corpus contains only the prior turn as  conversation context. Given that we are interested in studying also the succeeding turn as context, we checked to see whether for a current turn we can extract its succeeding turn from the general $IAC$ corpus. Out of the 4,692 current turns, we found that a total of 2,309 have a succeeding turn. We denote this corpus as $IAC^{+}_{v2}$. Since a candidate turn can have more than one succeeding reply in the $IAC$ corpus, the total size of the $IAC^{+}_{v2}$ dataset is 2,778. 
 Examples from the $IAC^{+}_{v2}$ are given in Table \ref{table:priorcurrentnext}.

\paragraph{{\bf Reddit Corpus}} \citet{khodak2017large} introduce the Self-Annotated Reddit Corpus (SARC) which is a very large collection of sarcastic and non-sarcastic posts  (over one million) from different subreddits. Similar to $IAC_{v2}$, this corpus also contains the prior turn as conversation context (the prior turn is either the original post or a prior turn in the discussion thread that the current turn is a reply to). Unlike $IAC_{v2}$, this corpus contains self-labeled data, that is the speakers labeled their posts/comments as sarcastic using the marker  ``/s'' to the end of sarcastic posts. For obvious reasons, the data is noisy since many users do not make use of the marker, do not know about it, or only use it where the sarcastic intent is not otherwise obvious. \citet{khodak2017large} have conducted an evaluation of the data having three human evaluators manually check a random subset of 500 comments from the corpus tagged as sarcastic and 500 tagged as non-sarcastic, with full access to the post's context. They found around 3\% of the non-sarcastic data is false negative. In their preliminary computational work on sarcasm detection, \citet{khodak2017large} have only selected posts between 2-50 words. For our research, we consider current turns that  contain several sentences (between three to seven sentences). 
We selected a subset of the corpus (a total of 50K instances balanced between both the categories). We will refer to this corpus as the $Reddit$ corpus. Table \ref{table:priorcurrent} shows an example of sarcastic current turn (userF's post) and its prior turn (userE's post) from the $Reddit$ dataset. We employ standard preprocessing, such as sentence boundary detection and word tokenization when necessary.\footnote{Unless stated otherwise, we use NLTK toolkit \citep{bird2009natural} for preprocessing.}


\paragraph{{\bf Twitter Corpus}} We have relied upon the annotations that users assign to their tweets using hashtags.  We used Twitter developer APIs to collect tweets for our research.\footnote{Particularly, we use teo libraries, the ``twitter4j'' (in Java) and the ``twarc'' (in Python) to accumulate the tweets.} 
The sarcastic tweets were collected using hashtags such as \emph{\#sarcasm}, \emph{\#sarcastic}, \emph{\#irony}. As non-sarcastic utterances, we consider sentiment tweets, i.e., we adopt the methodology proposed in related work \citep{gonzalez,muresanjasist2016}. The non-sarcastic tweets were the ones that do not contain the sarcasm hashtags, but are hashtags that contain positive or negative sentiment words. The positive tweets express direct positive sentiment and they are collected based on tweets with positive hashtags such as \emph{\#happy}, \emph{\#love}, \emph{\#lucky}. Similarly, the negative tweets express direct negative sentiment and are collected based on tweets with negative hashtags such as  \emph{\#sad}, \emph{\#hate}, \emph{\#angry}. Classifying sarcastic utterances against sentiment utterances is a considerably harder task than classifying against random objective tweets, since many sarcastic utterances also contain sentiment terms \citep{gonzalez,muresanjasist2016}. Table \ref{table:hashtags} shows all the hashtags used to collect the tweets. Similar to the $Reddit$ corpus, this is a self-labeled dataset, that is the speakers use \#hashtags to label their posts as sarcastic. We exclude retweets (i.e., tweets that start with ``RT''), duplicates, quotes and tweets that contain only hashtags and URLs or are shorter than three words. We also eliminated tweets in languages other than English using the library Textblob.\footnote{Textblob:http://textblob.readthedocs.io/en/dev/} Also, we eliminate all tweets where the hashtags of interest were not positioned at the very end of the message. Thus, we removed utterances such as ``\#sarcasm is something that I love''.

\begin{table}
\centering
\begin{small}
\begin{tabular}{p{3.5cm}|p{7.5cm}}
\hline
Type & Hashtags  \\
\hline
\textit{Sarcastic (S)} & \#sarcasm, \#sarcastic, \#irony \\
\textit{Non-Sarcastic (PosSent)} & \#happy, \#joy, \#happiness, \#love, \#grateful, \#optimistic, \#loved, \#excited, \#positive, \#wonderful, \#positivity, \#lucky \\
\textit{Non-Sarcastic (NegSent)} & \#angry, \#frustrated, \#sad, \#scared, \#awful, \#frustration, \#disappointed, \#fear, \#sadness, \#hate, \#stressed  \\
\hline
\end{tabular}
\caption{Hashtags for Collecting Sarcastic and Non-Sarcastic Tweets}
\label{table:hashtags}
\end{small}
\end{table}

To build the conversation context, for each sarcastic and non-sarcastic tweet we used the ``reply to status'' parameter in the tweet to determine whether it was in reply to a previous tweet: if so, we downloaded the last tweet (i.e., ``local conversation context'') to which the original tweet was replying to \citep{bamman2015contextualized}. In addition, we also collected the entire threaded conversation when available \citep{wang2015twitter}. Although we have collected over 200K tweets in the first step, around 13\% of them were a reply to another tweet, and thus our final Twitter conversations set contains 25,991 instances (12,215 instances for sarcastic class and 13,776 instances for the non-sarcastic class). We denote this dataset as the $Twitter$ dataset. We notice that 30\% of the tweets have more than one tweet in the conversation context. Table \ref{table:priorcurrent} shows an example of sarcastic current turn (userB's post) and its prior turn (userA's post) from the $Twitter$ dataset.  

There are two main differences between these datasets that need to be acknowledged. First, discussion forum posts are much longer than Twitter messages. Second, the way the gold labels for the sarcastic class are obtained is different. For the $IAC_{v2}$ and $IAC^{+}_{v2}$ datasets, the gold label is obtained via crowdsourcing, thus the gold label emphasizes whether the sarcastic intent is \emph{perceived} by the annotators (we do not know if the author intended to be sarcastic or not). For the $Twitter$ and the $Reddit$ datasets, the gold labels are given directly by the speakers (using hashtags on Twitter and the ``/s" marker in Reddit), signaling clearly the speaker's sarcastic intent. A third difference should be noted: the size of the $IAC_{v2}$ and $IAC^{+}_{v2}$ datasets is much smaller than the size of the $Twitter$ and $Reddit$ datasets. 

Table \ref{table:datanumbers} presents the size of the training, development, and test data for the four corpora. Table \ref{table:wordspost} presents the average number of words per post and the number of average sentences per post. The average number of words/post for the two discussion forums are comparable.   

\begin{table}
\centering
\begin{small}
\begin{tabular}{p{1.5cm}|p{3.5cm}|p{3.5cm}|p{3.5cm}}
\hline
Corpus & Train & Dev & Test  \\
\hline
$Twitter$ & 20,792 & 2,600 & 2,600  \\
$IAC_{v2}$ & 3,756 & 468 & 468 \\
$IAC^{+}_{v2}$  & 2,223 & 279 & 276 \\
$Reddit$ & 40,000 & 5,000 & 5,000 \\
\hline
\end{tabular}
\caption{Datasets description (number of instances in Train/Dev/Test)}
\label{table:datanumbers}
\end{small}
\end{table}

\begin{table}
\centering
\begin{small}
\begin{tabular}{c|c|c|c|c|c|c}

\hline
Corpus & \multicolumn{2}{c}{P\_{TURN}} & \multicolumn{2}{c}{C\_{TURN}} & \multicolumn{2}{c}{S\_{TURN}}  \\

& \#words & \#sents. & \#words & \#sents. & \#words & \#sents. \\
\hline
$Twitter$ & 17.48 & 1.71 & 16.92 & 1.51 & - & - \\
$IAC_{v2}$ & 57.22 & 3.18  & 52.94 & 3.50 & - & - \\
$IAC^{+}_{v2}$  & 47.94 & 2.55 & 42.82 & 2.98 & 43.48 & 3.04\\
$Reddit$ & 65.95 & 4.14 & 55.53 & 3.92 & - & -\\
\hline
\end{tabular}
\caption{Average words/post and sentences/post from the three corpora}
\label{table:wordspost}
\end{small}
\end{table}

\section{Computational Models and Experimental Setup} \label{section:experiment}

To answer the first research question ``does modeling of conversation context help in sarcasm detection"  we consider two binary classification tasks. We refer to sarcastic instances as $S$ and non-sarcastic instances as $NS$. 

The first task is to predict whether the current turn ({\sc c\_turn} abbreviated as $ct$) is sarcastic or not, considering it in isolation  --- $S^{ct}$ vs. $NS^{ct}$ task. 

The second task is to predict whether the current turn is sarcastic or not, considering both the current turn and its conversation context given by the prior turn ({\sc p\_turn}, abbreviated as $pt$), succeeding turn ({\sc s\_turn}, abbreviated as $st$)  or both ---  $S^{ct+context}$ vs. $NS^{ct+context}$ task, where  $context$ is $pt$, $st$ or $pt$+$st$. 

For all the corpora introduced in Section \ref{section:corpus} --- $IAC_{v2}$, $IAC^{+}_{v2}$, $Reddit$, and $Twitter$ --- we conduct $S^{ct}$ vs. $NS^{ct}$ and $S^{ct + pt}$ vs. $NS^{ct+pt}$ classification tasks. For $IAC^{+}_{v2}$ we also perform experiments considering the succeeding turn $st$ as conversation context (i.e., $S^{ct+st}$ vs. $NS^{ct+st}$ and $S^{ct+pt+st}$ vs. $NS^{ct+pt+st}$).  

We experiment with two types of computational models: (1) Support Vector Machines (SVM) \citep{CortesVapnik1995,libsvm} with linguistically-motivated discrete features (used as one baseline; disc$_{bl}$) and with tf-idf representations of the n-grams (used as another baseline; tf-idf$_{bl}$), and (2) approaches using distributed representations. For the latter, we use the Long Short-Term Memory (LSTM) Networks \citep{hochreiter1997long} that have been shown to be successful in various NLP tasks, such as constituency parsing \citep{vinyals2015grammar}, language modeling \citep{zaremba2014recurrent}, machine translation \citep{sutskever2014sequence} and textual entailment \citep{bowman2015large,rocktaschel2015reasoning,parikh2016decomposable}. We present these models in the next subsections.

\subsection{Baselines}  \label{subsection:discrete}

For features, we used n-grams, lexicon-based features, and sarcasm indicators that are commonly used in the existing sarcasm detection approaches \citep{tchokni2014,gonzalez,riloff,joshi2015,ghoshguomuresan2015EMNLP,muresanjasist2016,ghosh20181}. Below is a short description of the features.\\
\paragraph{{\bf BoW}} Features are derived from unigram, bigram, and trigram representation of words.\\

\paragraph{{\bf Lexicon-based features}} The lexicon-based features are derived from Pennebaker et al.'s Linguistic Inquiry and Word Count (LIWC) \citep{pennebaker2001} dictionary and emotion words from WordNet-Affect \citep{strapparava2004wordnet}. LIWC dictionary has been used widely in computational approaches to sarcasm detection \citep{gonzalez,muresanjasist2016,justo2014extracting}. It consists of a set of 64 word categories ranging from different \textit{Linguistic Processes} (e.g., Adverbs, Past Tense, Negation),  \textit{Psychological Processes} (e.g., Positive Emotions, Negative Emotions, Perceptual Processes [See, Hear, Feel], SocialProcesses); \textit{Personal Concerns}  (e.g., Work, Achievement, Leisure); and \textit{Spoken Categories} (Assent, Non-fluencies, Fillers). The LIWC dictionary contains around 4,500 words and word stems. Each category in this dictionary is treated as a separate feature, and we define a Boolean feature that indicates if a context or a reply contains a LIWC category. WordNet-Affect \citep{strapparava2004wordnet} is an affective lexical resource of words that extends WordNet by assigning a variety of affect labels to a subset of synsets representing affective concepts in WordNet. Similar to \citet{muresanjasist2016} we used the words annotated for associations with six emotions considered to be the most basic -- joy, sadness, fear, disgust, anger, and surprise \citep{ekman1992argument}, a total of 1,536 words. \\

\paragraph{{\bf Turn-Level Sentiment Features}} Two sentiment lexicons are also used to model the turn sentiment: the MPQA Sentiment Lexicon \cite{wilson2005recognizing} that contains over 8,000 positive, negative, and neutral sentiment words and an opinion lexicon that contains around 6,800 positive and negative sentiment words \citep{hu2004mining}. To capture sentiment at turn level, we count the number of positive and negative sentiment tokens, negations, and use a boolean feature that represents whether a turn contains both positive and negative sentiment tokens. For the $S^{ct+pt}$ vs. $NS^{ct+pt}$ classification task, we check whether the current turn $ct$ has a different sentiment than the prior turn $pt$ (similar to \citet{joshi2015}). Given that sarcastic utterances often contain a positive sentiment towards a negative situation, we hypothesize that this feature will capture this type of sentiment incongruity.\\

\paragraph{{\bf Sarcasm Markers}} \citet{burgers2012verbal} introduce a set of sarcasm markers that explicitly signal if an utterance is sarcastic. These markers are the meta-communicative clues that inform the reader that an utterance is sarcastic \citep{ghosh20181}. 
Three types of markers --- tropes (e.g., hyperbole), morpho-syntactic, and typographic are used as features. They are listed below.
\begin{enumerate}
\item \emph{Hyperbolic words:} Hyperboles or intensifiers are commonly used in sarcasm because speakers frequently overstate the magnitude of a situation or event. We use the MPQA lexicon \citep{wilson2005recognizing} to select hyperbolic words, i.e., words with very strong subjectivity. These words (e.g., ``greatest'', ``best'', ``nicest'' etc.) are common in ironic and sarcastic utterances \citep{camp2011}.
\item \emph{Morpho-syntactic:}
\begin{itemize}
\item \emph{Exclamations:} The use of exclamations (``!'') is standard in expressions of irony and sarcasm. They emphasize a sense of surprise on the literal evaluation that is reversed in the ironic reading \cite{burgers2010verbal}. Two binary features identify whether there is a single or multiple exclamation marks in the utterance. 
\item \emph{Tag Questions:} As shown in \citep{burgers2010verbal} tag questions are common in ironic utterances. We built a list of tag questions (e.g., ``didn't you?'', ``aren't we?'') from a grammar site and use them as binary indicators.\footnote{http://www.perfect-english-grammar.com/tag-questions.html} 
\item \emph{Interjections:} Interjections seem to undermine a literal evaluation and occur frequently in ironic utterances (e.g., ```yeah", `wow'', ``yay'',``ouch'',  etc.). Similar to tag questions we drew interjections (a total of 250) from different grammar sites. 
\end{itemize}
\item  \emph{Typography}:
\begin{itemize}
\item \emph{Capitalization:} Capitalization expresses excessive stress and thus it is standard in sarcastic posts on social media. For instance the words ``GREAT", ``SO" and ``WONDERFUL" are indicators of sarcasm in the example ``GREAT i'm SO happy;  shattered phone on this WONDERFUL day!!!''. 
\item \emph{Quotations:} This feature identifies whether any quotation appears in the utterance (i.e., replying to another message sarcastically).
\item \emph{Emoticons:} Emoticons are frequently used to emphasize the sarcastic intent of the user. In the example ``I love the weather ;) \#sarcasm'', the emoticon ``;)'' (wink) alerts the reader to a possible sarcastic interpretation. We collected  a comprehensive list of emoticons (over one-hundred) from Wikipedia and also used standard regular expressions to identify emoticons in our datasets.\footnote{http://sentiment.christopherpotts.net/code-data/} Besides using the emoticons directly as binary features, we use their sentiment as features as well (e.g., ``wink'' is regarded as positive sentiment in MPQA). \item \emph{Punctuations:}  Punctuation marks such as ``?'', ``.'', ``;'' and their various uses (e.g., single/multiple/mix of two different punctuations) are used as features.
\end{itemize}
\end{enumerate}

When building the features, we lowercased the utterances, except the words where all the characters are uppercased (i.e., we did not lowercase ``GREAT'', ``SO'', and ``WONDERFUL'' in the example above). Twitter tokenization is done by CMU's Tweeboparser \citep{gimpel2011part}. For the discussion forum dataset we use the NLTK tool \citep{bird2009natural} for sentence boundary detection and tokenization. We used the libSVM toolkit with Linear Kernel as well as RBF Kernel \citep{libsvm} with weights inversely proportional to the number of instances in each class. The SVM models build with these discrete-features will be one of our baselines (disc$_{bl}$). 

We also computed another baseline based on the tf-idf (i.e., term-frequency-inverse-document-frequency) features of the ngrams (e.g., unigrams, bigrams, and trigrams) from the respective turns and used SVM for the classification. The count of a candidate ngram in a turn is the term-frequency. The inverse document frequency is the logarithm of the division between total number of turns and number of turns with the ngram in the training dataset. This baseline is represented as tf-idf$_{bl}$ in the following sections.

\subsection{Long Short-Term Memory Networks}

LSTMs are a type of recurrent neural networks (RNNs) able to learn long-term dependencies \citep{hochreiter1997long}. Recently, LSTMs have been shown to be effective in Natural Language Inference (NLI) tasks such as Recognizing Textual Entailment, where the goal is to establish the \emph{relationship} between two inputs (e.g., a premise and a hypothesis) \citep{bowman2015large,rocktaschel2015reasoning,parikh2016decomposable}. LSTMs address the vanishing gradient problem commonly found in RNNs by incorporating gating functions into their state dynamics \citep{hochreiter1997long}. 
We introduce some notations and terminology standard in the LSTM literature \cite{tai2015}. The LSTM unit at each time step $t$ is defined as a collection of vectors: an input gate $i_t$, a forget gate $f_t$, an output gate $o_t$, a memory cell $c_t$, 
and a hidden state $h_t$. The LSTM transition equations are listed below: 

\begin{equation} \label{eq:lstm}
	\begin{split}
		&  i_t = \sigma(\mathbf W_i * [ h_{t-1},x_t] + b_i) \\
		& f_t = \sigma(\mathbf W_f * [ h_{t-1},x_t] + b_f) \\
		& o_t = \sigma(\mathbf W_o * [ h_{t-1},x_t] + b_o) \\
		&  \tilde{C}_t = \tanh(\mathbf W_c * [ h_{t-1},x_t] + b_c) \\
		& c_t = f_t \odot c_{t-1} +  i_t \odot \tilde{C}_{t}    \\
		&  h_t = o_t \odot \tanh(c_t) 
	\end{split}
\end{equation}

where $x_t$ is the input at the current time step, $\sigma$ is the logistic sigmoid function and $\odot$ denotes element-wise multiplication.  The input gate controls how much each unit is updated, the forget gate controls the extent to which the previous memory cell is forgotten, and the output gate controls the exposure of the internal memory state. The hidden state vector is a gated, partial view of the state of the unit's internal memory cell. Since the value of the gating variables vary for each vector element, the model can learn to represent information over multiple time scales.

As our goal is to explore the role of contextual information (e.g., prior turn and/or succeeding turn) for recognizing whether the current turn is sarcastic or not,  we will use \emph{multiple LSTMs}: one which reads the current turn and one (or two) which read(s) the context (e.g., one LSTM will read the prior turn and one will read the succeeding turn when available). 

\paragraph{{\bf Attention-based LSTM Networks}}
Attentive neural networks have been shown to perform well on a variety of NLP tasks \citep{yang2016hierarchical,yin2015abcnn,xu2015show}. Using attention-based LSTM will accomplish two goals: (1) test whether they achieve higher performance than simple LSTM models and (2) use the attention weights produced by the LSTM models to perform the qualitative analyses that enable us to answer the last two questions we want to address (e.g., which portions of context triggers the sarcastic reply). 

\citet{yang2016hierarchical} have included two levels of attention mechanisms, one at the word level and another at the sentence level where the sentences are in turn produced by attentions over words (i.e., the hierarchical model). We experiment with two architectures: one hierarchical that uses both word-level and sentence level attention \cite{yang2016hierarchical}, and one which uses only sentence-level attention (here we use only the average word embeddings to represent the sentences). 
One question we want to address is whether the sentence level attention weights indicate what sentence(s) in the prior turn trigger(s) the sarcastic reply. In the discussion forum datasets, prior turns are usually more than three sentences long and thus the attention weights could indicate what part of the prior turn triggers the sarcastic post $ct$. 

\begin{figure}[t]
\centering
\includegraphics[width=4.5in]{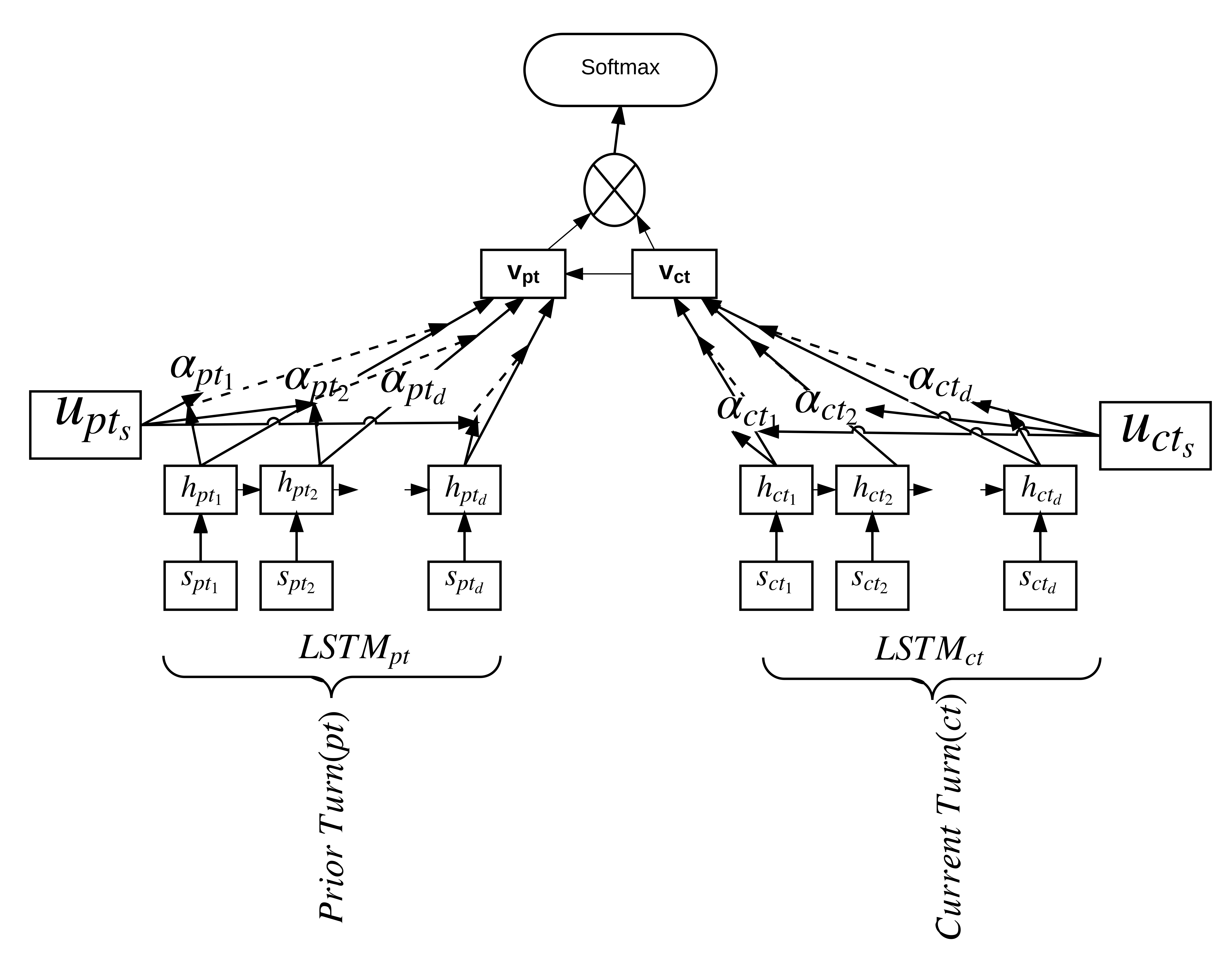}
\caption{Sentence-level Attention Network for prior turn $pt$ and current turn $ct$. Figure is inspired by \citet{yang2016hierarchical}}
\label{figure:model}
\end{figure}

Figure \ref{figure:model} shows the high-level structure of the model where the conversation context is represented by the prior turn $pt$. The context (left) is read by an LSTM ($LSTM_{pt}$) whereas the current turn $ct$ (right) is read by another LSTM ($LSTM_{ct}$). Note, for the model where we consider the succeeding turn $st$ as well, we simply use another LSTM to read $st$. For brevity, we only show the sentence-level attention.   

 Let the context $pt$ contain $d$ sentences and each sentence $s_{pt_{i}}$ contain $T_{pt_{i}}$ words. Similar to the notation of \citet{yang2016hierarchical}, we first feed the sentence annotation $h_{pt_{i}}$ through a one layer MLP to get $u_{pt_{i}}$ as a hidden representation of $h_{pt_{i}}$, then we weight the sentence $u_{pt_{i}}$ by measuring similarity with a sentence level context vector $u_{{pt}_{s}}$. This gives a normalized importance weight $\alpha_{pt_{i}}$ through a softmax function. $v_{pt}$ is the vector that summarize all the information of sentences in the context ($LSTM_{pt}$).
\begin{equation} \label{eq:attention}
v_{pt} = \sum_{i\in[1,d]} \alpha_{{pt}_{i}}h_{{pt}_{i}}
\end{equation}
where attention is calculated as:
\begin{equation}
\alpha_{{pt}_{i}} = \frac{\exp(u_{{pt}_{i}}^Tu_{{pt}_{s}} )}{\sum_{i\in[1,d]}\exp(u_{{pt}_{i}}^Tu_{{pt}_{s}} )}
\end{equation}

Likewise we compute $v_{ct}$ for the current turn $ct$ via $LSTM_{ct}$ (similar to equation \ref{eq:attention}; also shown in Figure \ref{figure:model}). Finally, we concatenate the vector $v_{pt}$ and $v_{ct}$ from the two LSTMs for the final softmax decision (i.e., predicting the $S$ or $NS$ class). In case of using the succeeding turn $st$ also in the model, we concatenate the vectors $v_{pt}$, $v_{ct}$ and $v_{st}$.      

As stated earlier in this section, we also experiment with both word and sentence level attentions in a hierarchical fashion similarly to the approach proposed by \citet{yang2016hierarchical}. As we show in Section \ref{section:results}, however, we achieve the best performance using just the sentence-level attention. A possible explanation is that attention over both words and sentences seek to learn a large number of model parameters and given the moderate size of the discussion forum corpora they might overfit.

For tweets, we treat each individual tweet as a sentence. The majority of tweets consist of a single sentence and even if there are multiple sentences in a tweet, often one sentence contains only hashtags, URLs, and emoticons making them uninformative if treated in isolation. 

\begin{figure}[t]
\centering
\includegraphics[width=4.5in]{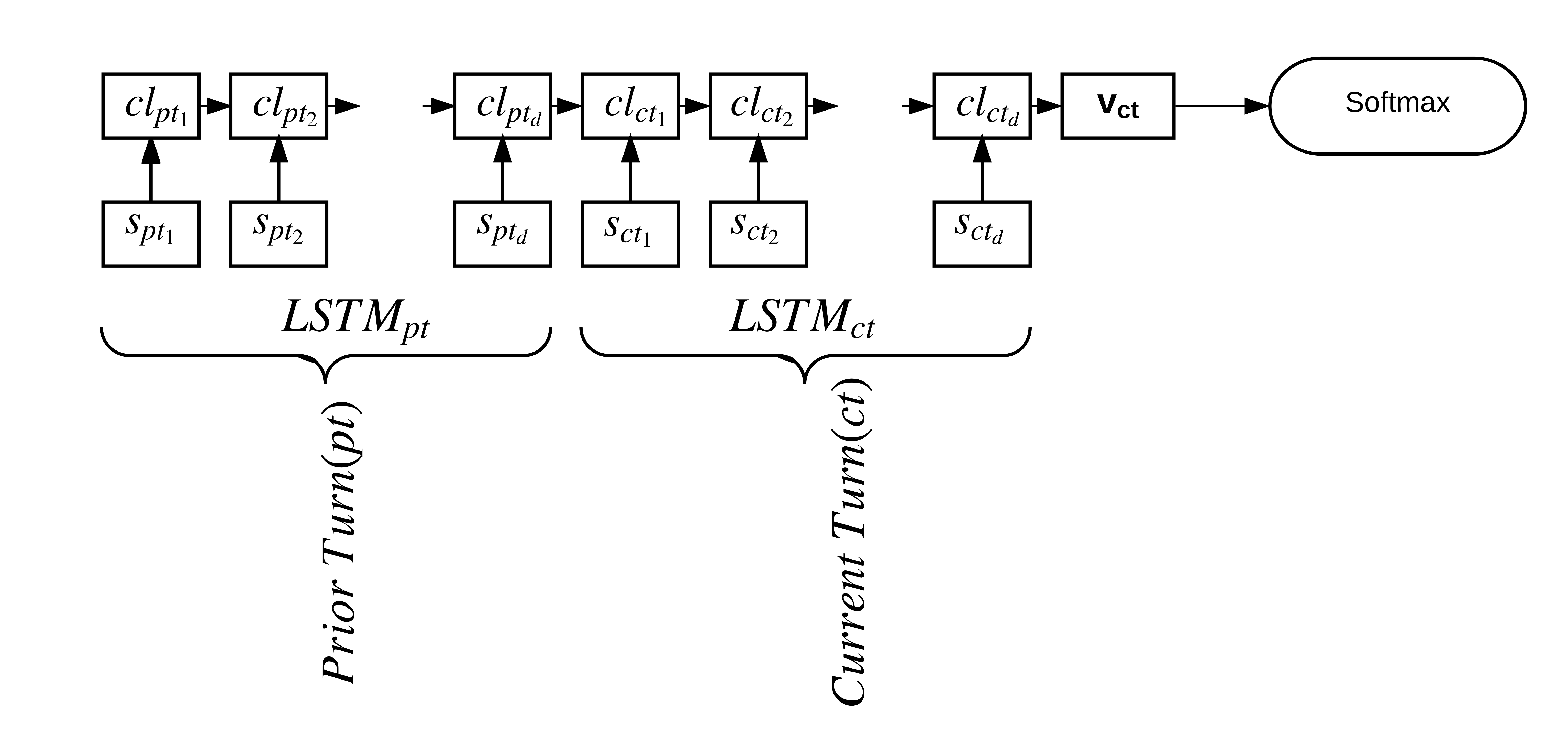}
\caption{Conditional LSTM Network for prior turn $pt$ and current turn $ct$; Figure is inspired by the model proposed in \citet{rocktaschel2015reasoning}}
\label{figure:conditional}
\end{figure}

\paragraph{{\bf Conditional LSTM Networks}}
We also experiment with the \emph{conditional encoding} model as introduced by  \citet{rocktaschel2015reasoning} for the task of recognizing textual entailment. In this architecture,  two separate LSTMs are used -- $LSTM_{pt}$ and $LSTM_{ct}$ -- similar to the previous architecture without any attention, but for $LSTM_{ct}$, its memory state is initialized with the last cell state of $LSTM_{pt}$. In other words, $LSTM_{ct}$ is conditioned on the representation of the $LSTM_{pt}$ that is built on the prior turn $pt$. For models that use the successive turn $st$ as the context the LSTM representation $LSTM_{st}$ is conditioned on the representation of the $LSTM_{ct}$. Figure \ref{figure:conditional} shows the LSTM network where the current turn $ct$ is conditioned on the prior turn $pt$.    

\paragraph{{\bf Parameters and pre-trained word vectors}}
All datasets were split randomly into training (80\%), development (10\%), and test (10\%), maintaining the same distribution of sarcastic vs. non-sarcastic classes. For Twitter, we used the skip-gram word-embeddings (100-dimension) used in \citep{ghoshguomuresan2015EMNLP} that was built using over 2.5 million tweets.\footnote{https://github.com/debanjanghosh/sarcasm\_wsd} For discussion forums, we use the standard Google n-gram $word2vec$ pre-trained model (300-dimension) \citep{mikolov2013efficient}. Out-of-vocabulary words in the training set are randomly initialized via sampling values uniformly from (-0.05, 0.05) and optimized during training. We use the development data to tune the parameters (e.g., dropout rate, batch-size, number of epochs, $L2$-regularization) and selected dropout rate \citep{srivastava2014dropout} of 0.5 (from [.25, 0.5, 0.75]), mini-batch size of 16, $L2$-regularization to 1E-4 (from [1E-2, 1E-3, 1E-4, 1E-5]) and set the number of epochs to 30. We set the threshold of the maximum number of sentences to ten per post so that for any post that is longer than ten sentences we select the first ten sentences for our experiments. Finally, the maximum number of words per sentence is set at fifty (zero-padding is used when necessary).  

\section{Results} \label{section:results} 

In this section, we present a quantitative analysis aimed at addressing our first question ``does modeling conversation context help in sarcasm detection?" First, we consider just the \emph{prior turn as conversation context} and show results of our various models on all datasets: $IAC_{v2}$, $Reddit$, and $Twitter$ (Section \ref{section:pt}). Also, we perform an experiment where we train on $Reddit$ (discussion forum, self-labeled) and test on $IAC_{v2}$ (discussion forum, labeled via crowdsourcing). Second, we consider \emph{both the prior turn and the succeeding turn as context} and report results of various models on our $IAC^{+}_{v2}$ dataset (Section \ref{section:st}). We report Precision (P), Recall (R), and F1 scores on sarcastic ($S$) and non-sarcastic ($NS$) classes. We conclude the section with an error analysis of our models (Section \ref{section:ea}). 

\subsection{Prior Turn as Conversation Context} \label{section:pt}

We use two baselines, depicted as $bl$. First, disc$_{bl}^{ct}$ and disc$_{bl}^{ct+pt}$ represent the performance of the SVM models with discrete features (Section \ref{subsection:discrete}) when using only the current turn $ct$ and the $ct$ together with the prior turn $pt$, respectively. Second is the tf-idf based baseline. Here, tf-idf$_{bl}^{ct}$ and tf-idf$_{bl}^{ct+pt}$ represent the performance of tf-idf values of $ct$ and the $ct$ together with the prior turn $pt$, respectively. 

We experimented with both linear and RBF kernel and observed that linear kernel consistently performed better than the RBF kernel. Only in the case of $IAC_{v2}$, for tf-idf$_{bl}^{ct+pt}$ setting, RBK kernel performed better (68.15\% F1 for category $S$ and 64.56\% F1 for category $NS$). Thus,  we only report the performance of the linear kernel for all the experiments. 

Although we did not apply any feature selection, we use frequency threshold to select the n-grams (the minimum count is 5). Likewise, for the tf-idf based representation, the minimum frequency (i.e., DF) is set to 5. We use the development data to empirically select this minimum frequency. We also use the standard stop word list provided by the NLTK toolkit.   

LSTM$^{ct}$ and LSTM$^{ct+pt}$ represent the performance of the simple LSTM models when using only the current turn $ct$ and the $ct$ concatenated with the prior turn $pt$, respectively. LSTM$^{ct}$ + LSTM$^{pt}$ depicts the multiple-LSTM model where one LSTM is applied on the current turn, ${ct}$ and the other LSTM is applied on the prior turn, ${pt}$. LSTM$^{pt_{a}}$ and LSTM$^{ct_{a}}$ are the attention-based LSTM models of context $pt$ and current turn $ct$, where the $w$, $s$ and $w+s$ subscripts denote the word-level, sentence-level or word and sentence level attentions. LSTM$^{conditional}$ is the \emph{conditional encoding} model that conditions the LSTM that reads the current turn on the LSTM that reads the prior turn (no attention). 
Given these notations, we present the results on each of the three datasets. 

\paragraph{{\bf $IAC_{v2}$} Corpus} Table \ref{table:results2} shows the classification results on the $IAC_{v2}$ dataset. Although a vast majority of the prior turn posts contain 3-4 sentences, around 100 have more than ten sentences and thus we set a cutoff to a maximum of ten sentences, for context modeling. For the current turn $ct$, we consider the entire post.
\begin{table*} [t]
\centering
\begin{tabular}{lcccccc}
\hline
\multirow {2}{*}{Experiment} & \multicolumn{3}{c}{$S$} &  \multicolumn{3}{c}{$NS$} \\
& P & R & F1 & P & R & F1  \\
\hline
{disc$_{bl}^{ct}$}  & 65.55  & 66.67  &  66.10   &  66.10  &  64.96 &  65.52   \\
{dics$_{bl}^{ct+pt}$}  & 63.32  & 61.97  &  62.63   &  62.77  &  64.10 &  63.50   \\
{tf-idf$_{bl}^{ct}$}  & 66.07  & 63.25  &  64.63   &  64.75  &  67.52 &  66.11   \\

{tf-idf$_{bl}^{ct+pt}$}  & 63.95  & 63.68  &  63.81   &  63.83  &  64.10 &  63.97  \\

{LSTM$^{ct}$}  & 67.90  & 66.23   & 67.10  & 67.08   & \textbf{68.80}  & 67.93    \\

{LSTM$^{ct+pt}$}  & 65.16  & 67.95   & 66.53  & 66.52   & {63.68}  & 65.07    \\


{LSTM$^{ct}$+LSTM$^{pt}$}  & 66.19 & 79.49 & 72.23 & 74.33 & 59.40 & 66.03  \\
{LSTM$^{conditional}$} & \textbf{70.03} & 76.92 & \textbf{73.32} &  74.41 & 67.10 & \textbf{70.56} \\ 
{LSTM$^{ct_{a_{s}}}$}  & 69.45 & {70.94} &  {70.19} & {70.30} & 68.80 & 69.45 \\
{LSTM$^{ct_{a_{s}}+{pt_{a_{s}}}}$}  & 64.46 & {69.33} &  {66.81} & {67.61} & 62.61 & 65.01 \\


{LSTM$^{ct_{a_{s}}}$+LSTM$^{pt_{a_{s}}}$}  & 66.90 & \textbf{82.05} & \textbf{73.70} & \textbf{76.80} & 59.40 & 66.99 \\
{LSTM$^{ct_{a_{w+s}}}$+LSTM$^{pt_{a_{w+s}}}$}  & 65.90 & 74.35 & 69.88 & 70.59 & 61.53 & 65.75 \\
\end{tabular}
\caption{Experimental results for the discussion forum dataset ($IAC_{v2}$) ({\bf bold} are best scores)}
\centering
\label{table:results2}
\end{table*}

We observe that both the baseline models, $disc_{bl}$ that is based on discrete features and $tf-idf_{bl}$ that is based on tf-idf values did not perform very well, and adding the context of the prior turn $pt$ actually hurt the performance. 
Regarding the performance of the neural network models, we observed that the multiple-LSTMs  model (one LSTM reads the context ($pt$) and  one reads the current turn($ct$), LSTM$^{ct}$ + LSTM$^{pt}$,  outperforms the model using just the current turn (results are statistically significant when compared to LSTM$^{ct}$). On the other hand, if only using one LSTM to model both prior turn and current turn (LSTM$^{ct+pt}$), it does not improve over just using the current turn, and has lower performance than the multiple-LSTM model (the results apply to the attention models as well). 
The highest performance when considering both the $S$ and $NS$ classes is achieved by the LSTM$^{conditional}$ model (73.32\% F1 for $S$ class and 70.56\% F1 for $NS$, showing a 6\% and 3\% improvement over LSTM$^{ct}$ for $S$ and $NS$ classes, respectively). The LSTM model with sentence-level attentions on both context and current turn ({LSTM$^{ct_{a_{s}}}$+LSTM$^{pt_{a_{s}}}$}) gives the best F1 score of 73.7\% for the $S$ class. For the $NS$ class, while we notice an improvement in precision we also notice a drop in recall when compared to the LSTM model with sentence-level attention only on the current post (LSTM$^{ct_{a_{s}}}$).  Remember that sentence-level attentions are based on average word embeddings. We also experimented with the hierarchical attention model where each sentence is represented by a \textit{weighted average} of its word embeddings. In this case, attentions are based on words and sentences, and we follow the architecture of hierarchical attention network \citep{yang2016hierarchical}. We observe that the performance (69.88\% F1 for $S$ category) deteriorates, probably due to the lack of enough training data. Since attention over both the words and sentences seek to learn more model parameters, adding more training data will be helpful. For the $Reddit$ and $Twitter$ data (see below), these models become better, but still not on par with just sentence-level attention showing that even larger datasets might be needed. 


\paragraph{{\bf $Twitter$} Corpus} Table \ref{table:results1} shows the results of the Twitter dataset. As with the $IAC_{v2}$ dataset, adding context using the discrete as well as the tf-idf features do not show a statistically significant improvement.  
For the neural networks models, similar to the results on the $IAC_{v2}$ dataset, the LSTM models that read both the context and the current turn outperform the LSTM model that reads only the current turn (LSTM$^{ct}$). However, unlike the $IAC_{v2}$ corpus, 
for Twitter, we observe that for the LSTM without attention, the single LSTM architecture (i.e., LSTM$^{ct+pt}$)  performs better, i.e., 72\% F1 between the sarcastic and non-sarcastic category (average) that is around 4\% better than the multiple LSTMs (i.e., LSTM$^{ct}$+LSTM$^{pt}$). Since tweets are short texts, often the prior or the current turns are only a couple of words, hence concatenating the prior turn and current turn would give more context to the LSTM model. However, for sentence-level attention models, multiple-LSTM are still a better choice than using a single LSTM and concatenating the context and current turn. 
The best performing architectures are again the LSTM$^{conditional}$ and LSTM with sentence-level attentions ({LSTM$^{ct_{a_{s}}}$+LSTM$^{pt_{a_{s}}}$}). LSTM$^{conditional}$ model shows an improvement of 11\% F1 on the $S$ class and 4-5\%F1 on the $NS$ class, compared to LSTM$^{ct}$. For the attention-based models, the improvement using context is smaller ($\sim$2\% F1). We kept the maximum length of prior tweets to the last five tweets in the conversation context, when available. We also considered an experiment with only the ``last'' tweet (i.e., {LSTM$^{ct_{a_{s}}}$+LSTM$^{last\_pt_{a_{s}}}$}), i.e., considering only the ``local conversation context'' (See Section \ref{section:corpus}). We observe that although the F1 for the non-sarcastic category is high (76\%), for the sarcastic category it is low (e.g., 71.3\%). This shows that considering a larger conversation context of multiple prior turns rather than just the last prior turn could assist in achieving higher accuracy particularly in Twitter where each turn/tweet is short.  
\begin{table*} [t] 
\centering
\begin{tabular}{lcccccc}
\hline
\multirow {2}{*}{Experiment} & \multicolumn{3}{c}{$S$} &  \multicolumn{3}{c}{$NS$} \\
& P & R & F1 & P & R & F1  \\
\hline
{disc$_{bl}^{ct}$} &	64.20	&	64.95	&	64.57	&	69.0	&	68.30	&	68.70 \\
{disc$_{bl}^{ct+pt}$} & 	65.64&		65.86 &		65.75	&	70.11 & 69.91	&	70.00 \\
{tf-idf$_{bl}^{ct}$} &	63.16	&	67.94	&	65.46	&	70.04	&	65.41	&	67.64 \\
{tf-idf$_{bl}^{ct+pt}$} & 	65.54&		72.86 &		69.01	&	73.75 & 66.57	&	69.98\\

{LSTM$^{ct}$}	& 73.25	&	58.72	&	65.19	&	61.47	&	75.44	&	67.74 \\

{LSTM$^{ct+pt}$}	& 70.54	&	71.19	&	70.80	&	64.65	&	74.06	&	\textbf{74.35} \\


{LSTM$^{ct}$+LSTM$^{pt}$}	 & 70.89	&	67.95	&	69.39 &		64.94	 &	68.03	&	66.45 \\ 
{LSTM$^{conditional}$} &	76.08	 &	\textbf{76.53}	&	\textbf{76.30}	&	\textbf{72.93}	&	72.44		& {72.68}  \\
{LSTM$^{ct_{a_{s}}}$} &	{76.00}	&	73.18	&	{74.56} 	&	70.52	&	{73.52}	&	{71.90} \\

{LSTM$^{ct_{a_{s}}+pt_{a_{s}}}$} &	{70.44}	&	67.28	&	{68.82} 	&	72.52	&	{75.36}	&	\textbf{73.91} \\


{LSTM$^{ct_{a_{s}}}$+LSTM$^{pt_{a_{s}}}$} &	\textbf{77.25}	&	75.51	&	\textbf{76.36} 	&	72.65	&	\textbf{74.52}	&	\textbf{73.57} \\
{LSTM$^{ct_{a_{s}}}$+LSTM$^{last\_pt_{a_{s}}}$} &	{73.10}	&	69.69	&	{71.36} 	&	74.58	&	\textbf{77.62}	&	\textbf{76.07} \\
{LSTM$^{ct_{a_{w}}}$+LSTM$^{pt_{a_{w}}}$} &	76.74	&	69.77	&	73.09	&	68.63	&	75.77	&	72.02 \\
{LSTM$^{ct_{a_{w+s}}}$+LSTM$^{pt_{a_{w+s}}}$} 	& 76.42	&	71.37	&	73.81	&	69.50	&	74.77	&	72.04 \\

\end{tabular}
\caption{Experimental results for Twitter dataset ({\bf bold} are best scores)}
\label{table:results1}
\end{table*}

\paragraph{Reddit Corpus}
Table \ref{table:resultsreddit} shows the results of the experiments on Reddit data. There are two major differences between this corpus and the $IAC_{v2}$ corpus. First, since the original release of the Reddit corpus \citep{khodak2017large} is very large, we select a subcorpus that is much larger than the $IAC_{v2}$ data containing 50K instances. In addition, we 
selected posts (both $pt$ and $ct$) that consist of a maximum of seven sentences primarily to be comparable with the $IAC_{v2}$ data.\footnote{$IAC_{v2}$ contains prior and current turns which contains mostly seven or fewer sentences.} Second, unlike the $IAC_{v2}$ corpus, the sarcastic current turns $ct$ are self-labeled, so it is unknown whether there are any similarities between the nature of the data in the two discussion forums. 

We observe that the baseline models (e.g., discrete as well as tf-idf features) perform similarly to the other discussion forum corpus $IAC_{v2}$. The {disc$_{bl}^{ct+pt}$} model performs poorly compared to the disc$_{bl}^{ct}$ model. Note, \citet{khodak2017large} have evaluated the sarcastic utterances via BoW features and sentence embeddings and achieved accuracy in mid 70\%. However, they selected sentences between two and fifty words for the classification which is very different from our setups, where we use larger comments (up to seven sentences).

Similar to the  $IAC_{v2}$ corpus, we observed that the multiple-LSTMs  models (one LSTM reads the context ($pt$) and  one reads the current turn($ct$)),  outperform the models using just the current turn (results are statistically significant both for simple LSTM and LSTM with attentions). Multiple-LSTM with sentence-level attention performs best. Using one LSTM to model both prior turn and current turn  has lower performance than the multiple-LSTM models. 

We also conducted experiments with word and sentence-level attentions (i.e., LSTM$^{ct_{a_{w+s}}}$+LSTM$^{pt_{a_{w+s}}}$). Even though we obtain slightly lower accuracy (i.e., 76.67\% for the sarcastic category) in comparison to sentence-level attention models, the difference is not as high as for the other corpora, which we believe is due to the larger size of the training data.

\begin{table*} [t] 
\centering
\begin{tabular}{lcccccc}
\hline
\multirow {2}{*}{Experiment} & \multicolumn{3}{c}{$S$} &  \multicolumn{3}{c}{$NS$} \\
& P & R & F1 & P & R & F1  \\
\hline
{disc$_{bl}^{ct}$} &	72.54	&	72.92 &	72.73	&	72.77	&	72.4	&	72.56 \\
{disc$_{bl}^{ct+pt}$} & 	66.3 &		67.52 &		66.90	&	66.91 & 65.68	&	66.29 \\
{tf-idf$_{bl}^{ct}$} &	72.76	&	70.08 &	71.39	&	71.14	&	73.76	&	72.43 \\

{tf-idf$_{bl}^{ct+pt}$} & 	71.14 &		69.72 &		70.42	&	70.31 & 71.72	& 71.01 \\

{LSTM$^{ct}$}	& {\bf 81.29} &   59.6 &   68.77 	&	68.1 & {\bf 86.28} &  {\bf 76.12} \\

{LSTM$^{ct+pt}$}	& { 73.35} &   75.76 &   74.54	&	74.94 & { 72.48} &  { 73.69} \\


{LSTM$^{ct}$+LSTM$^{pt}$}	 & 74.46 &  73.72 &  74.09  & 73.98 & 74.72 &  74.35 \\ 
{LSTM$^{conditional}$} &	73.72	 &	71.6	&	72.64	&	72.40	&	74.48		& 73.42  \\
{LSTM$^{ct_{a_{s}}}$} &	74.87 & 74.28 &  74.58 	&	74.48 &   75.08 &   {\bf 74.78} \\

{LSTM$^{ct_{a_{s}}+pt_{a_{s}}}$} &	77.24 & 69.83 &  73.35 	&	72.66 &   79.58 &   {\bf 75.96} \\


{LSTM$^{ct_{a_{s}}}$+LSTM$^{pt_{a_{s}}}$} &	73.11 &   {\bf 80.60} & {\bf 76.67} &  { \bf 78.39} &  70.36 &  74.16  \\

{LSTM$^{ct_{a_{w+s}}}$+LSTM$^{pt_{a_{w+s}}}$} 	& 74.50	&	74.68	&	{\bf 74.59}	&	74.62	&	74.44	&	74.52 \\

\end{tabular}
\caption{Experimental results for $Reddit$ dataset ({\bf bold} are best scores)}
\label{table:resultsreddit}
\end{table*}

\paragraph{Impact of the size and nature of the corpus}
Overall, while the results on the $Reddit$ dataset are slightly better than on the $IAC_{v2}$, given that the $Reddit$ corpus is ten times larger, we believe that the self-labeled nature of the $Reddit$ dataset might make the problem harder. To verify this hypothesis, we conducted two separate experiments. First, we selected a subset of the $Reddit$ corpus that is equivalent to the $IAC_{v2}$ corpus size (i.e., 5,000 examples balanced between the sarcastic and the not-sarcastic categories). We use the best LSTM model (i.e., attention on prior and current turn) which achieves 69.17\% and 71.54\% F1 for the sarcastic ($S$) and the non sarcastic ($NS$) class, respectively. These results  are lower than the ones we obtained for the $IAC_{v2}$ corpus using the same amount of training data and much lower than the performances reported on Table \ref{table:resultsreddit}. Second, we conducted an experiment where we trained our best models (i.e., LSTM models with sentence-level attention) on the $Reddit$ corpus and tested on the test portion of the $IAC_{v2}$ corpus. The results, shown in Table \ref{table:cross}, are much lower than when training using ten times less amount of data from $IAC_{v2}$ corpus, particularly for the sarcastic class (more than 10\% F1 measure drop). Moreover, unlike all the experiments, adding context does not help the classifier which seems to highlight a difference between the nature of the two datasets including the gold annotations (self-labeled for $Reddit$  vs. crowdsource labeled for $IAC_{v2}$) and most likely the topics covered by these discussion forums. 


\begin{table*} [t] 
\centering
\begin{tabular}{lcccccc}
\hline
\multirow {2}{*}{Experiment} & \multicolumn{3}{c}{$S$} &  \multicolumn{3}{c}{$NS$} \\
& P & R & F1 & P & R & F1  \\
\hline
{LSTM$^{ct_{a_{s}}}$} & 66.51	 &  61.11 &  63.69	&	64.03  & 69.23   & 66.53    \\
{LSTM$^{ct_{a_{s}}}$+LSTM$^{pt_{a_{s}}}$} &	 63.96 &	60.68&	62.28	&62.60	&65.81	& 64.17  \\
\end{tabular}
\caption{Experimental results for training on $Reddit$ dataset and testing on $IAC_{v2}$ using the best LSTM models (sentence-level attention).}
\label{table:cross}
\end{table*}



\paragraph{Impact of unbalanced datasets}

In previous experiments we used a balanced data scenario. However, in online conversations we are most likely faced with an unbalanced problem (the sarcastic class is more rare than the non-sarcastic class). We thus experimented with an unbalanced setting, where we have more instances of the non-sarcastic class ($NS$) than sarcastic class ($S$)(e.g., 2-times, 3-times, 4-times more data). We observe that the performance drops for the $S$ category in the unbalanced settings as expected. Table \ref{table:imbalance} shows the results of the unbalanced setting; particularly we show the setting where the $NS$ category has 4-times more training instances than the $S$ category. We used the $Reddit$ dataset since it had a larger amount of examples. For this experiments we used the best model from the balanced data scenario which was the LSTM with sentence level attention. In general, we observe that the Recall of the $S$ category is low and that impacts the F1 score.  During the LSTM training,  class weights (inversely proportional to the sample sizes for each class) are added to the loss function to handle the unbalanced data scenario. 
We observe that adding contextual information (i.e., LSTM$^{ct_{a_{s}}}$+LSTM$^{pt_{a_{s}}}$) helps the LSTM model and that pushes the F1 to 45\% (i.e., a six point improvement over LSTM$^{ct_{a_{s}}}$).

\begin{table*} [t] 
\centering
\begin{tabular}{lcccccc}
\hline
\multirow {2}{*}{Experiment} & \multicolumn{3}{c}{$S$} &  \multicolumn{3}{c}{$NS$} \\
& P & R & F1 & P & R & F1  \\
\hline
{LSTM$^{ct_{a_{s}}}$}	& 67.08 &    27.50 &   39.00 	&	84.32 &  96.66 &  90.07 \\
{LSTM$^{ct_{a_{s}}}$+LSTM$^{pt_{a_{s}}}$}	 & 62.25 &  35.05 &  44.85  &  85.48 &  94.73 &  89.87 \\ 


\end{tabular}
\caption{Experimental results for $Reddit$ dataset under unbalanced setting}
\label{table:imbalance}
\end{table*}

\subsection{Prior Turn and Subsequent Turn as Conversation Context} \label{section:st}

We also experiment using both the prior turn $pt$ and the succeeding turn $st$ as conversation context. Table \ref{table:resultsv2st} shows the experiments on the $IAC^{+}_{{v2}}$ corpus. We observe that the performance of the LSTM models is high in general (i.e., F1 scores in between 78-84\%, consistently for both the sarcastic ($S$) and non-sarcastic ($NS$) classes) compared to the discrete feature based models (i.e., disc$_{bl}$).   Table \ref{table:resultsv2st} shows that when we use conversation context, particularly the prior turn $pt$ or the prior turn and the succeeding turn together, the performance improves (i.e., around 3\% F1 improvement for sarcastic category and almost 6\% F1 improvement for non-sarcastic category). For the $S$ category, the highest F1 is achieved by the LSTM$^{ct}$+LSTM$^{pt}$ model (i.e., 83.92\%) whereas the LSTM$^{ct}$+LSTM$^{pt}$+LSTM$^{st}$ model performs best for the non-sarcastic class (83.09\%). Here, in the case of concatenating the turns and using a single LSTM (i.e., LSTM$^{ct+pt+st}$), the average F1 between the sarcastic and non-sarcastic category is 80.8\%, which is around 3.5\% lower than using separate LSTMs for separate turns ((LSTM$^{ct}$+LSTM$^{pt}$+LSTM$^{st}$). In comparison to the attention-based models, although using attention over prior turn $pt$ and successive turn $st$ helps in sarcasm identification compared to the attention over only the current turn $ct$ (i.e., improvement of around 2\% F1 for both the sarcastic as well as the non-sarcastic class), generally the accuracy is slightly lower than the models without attention. We suspect this is due to the small size of the $IAC^{+}_{{v2}}$ corpus ($<$ 3,000 instances). 

We also observe that the numbers obtained for $IAC^{+}_{{v2}}$ are higher than the one for the $IAC_{v2}$ corpus even if less training data is used. To understand the difference, we analyze the type of the sarcastic and non-sarcastic posts from the $IAC^{+}_{{v2}}$ and found that almost 94\% of the corpus consists of sarcastic messages of ``general'' type, 5\% of ``rhetorical questions ($RQ$)" type and very few (0.6\%) examples of the ``hyperbolic" type \citep{orabycreating}. Looking at \citet{orabycreating} it seems the ``general" type obtains the best results (Table 7 in \citet{orabycreating}) with almost 10\% F1 over the ``hyperbolic" type. As we stated before, although $IAC_{v2}$ corpus is larger than the $IAC^{+}_{{v2}}$ corpus, $IAC_{v2}$ maintains exactly the same distribution of ``general", ``$RQ$", and ``hyperbolic" examples. This also explains why Table \ref{table:resultsv2st} shows superior results since classifying the ``generic" type of sarcasm could be an easier task.

\begin{table*} [t] 
\centering
\begin{tabular}{lcccccc}
\hline
\multirow {2}{*}{Experiment} & \multicolumn{3}{c}{$S$} &  \multicolumn{3}{c}{$NS$} \\
& P & R & F1 & P & R & F1  \\
\hline
{disc$_{bl}^{ct}$} &	76.97	&	78.67 &	77.81	&	78.83	&	77.14	&	77.97 \\
{disc$_{bl}^{ct+pt}$} & 	76.69 &		75.0&		75.83	&	76.22 & 77.85	&	77.03 \\
{disc$_{bl}^{ct+st}$} & 	67.36 &		71.32 &		69.28	&	70.45 & 66.43	&	68.38 \\
{disc$_{bl}^{ct+pt+st}$} & 	74.02 &		69.12 &		71.48	&	71.81 & 76.43	&	74.05 \\
{tf-idf$_{bl}^{ct}$} &	71.97	&	69.85 &	70.90 &	71.53	&	73.57	&	72.54 \\

{tf-idf$_{bl}^{ct+pt}$} & 	72.66 &		74.26&		73.45	&	74.45 & 72.86	&	73.65 \\

{tf-idf$_{bl}^{ct+st}$} & 	72.73 &		70.59 &		71.64	&	72.22 & 74.29	&	73.24 \\

{tf-idf$_{bl}^{ct+pt+st}$} & 	75.97 &		72.06 &		73.96	&	74.15 & 77.86 &	75.96 \\

{LSTM$^{ct}$}	& 74.84 &    87.50 &   80.68 	&	85.47 &  71.43 &  77.82 \\
{LSTM$^{ct+pt}$}	 & { 69.03} &  78.67 & {  73.53 } & 76.03 &  65.71 &  {70.49}   \\ 
{LSTM$^{ct+st}$}	 & { 78.38} &  85.29 & {  81.60 } & 84.37 &  77.14 &  {80.59}   \\ 
{LSTM$^{ct+pt+st}$}	 & { 76.62} & 88.06 & {  81.94 } & 86.55 &  74.10 &  {79.84}   \\ 
{LSTM$^{ct}$+LSTM$^{pt}$}	 & 80.00 &  {\bf 88.24} &  {\bf  83.92}  & {\bf 87.30} &  78.57 &  {\bf 82.71} \\ 
{LSTM$^{ct}$+LSTM$^{st}$}	 & 79.73 &   86.76 &   83.10  & 85.94 &  78.57 &  82.09 \\ 
{LSTM$^{ct}$+LSTM$^{pt}$+LSTM$^{st}$}	 & {\bf 81.25} &  86.03 & {\bf  83.57 } & 85.61 &  80.71 &  {\bf 83.09}   \\ 
{LSTM$^{conditional(pt->ct)}$} &	79.26	 &	{78.68}	&	{78.97}	&	{79.43}	&	80.00		& {79.71}  \\
{LSTM$^{conditional(ct->st)}$} &	70.89	 &	69.85	&	70.37	&	{71.13}	&	72.14		& {71.63} \\
{LSTM$^{ct_{a_{s}}}$} &	77.18 &  84.56 &   80.70 	&	83.46 &   75.71 &   79.40 \\
{LSTM$^{ct_{a_{s}}}$+LSTM$^{pt_{a_{s}}}$} &	{\bf 80.14} &   83.09 &  81.59 &   82.96 &  {\bf 80.00} &  81.45  \\
{LSTM$^{ct_{a_{s}}}$+LSTM$^{st_{a_{s}}}$} &	75.78 &  {\bf 89.71} &  82.15 &    {\bf 87.83}  &  72.14 &  79.22 \\

{LSTM$^{ct_{a_{s}}}$+LSTM$^{pt_{a_{s}}}$+LSTM$^{st_{a_{s}}}$} &	76.58 &   {\bf 88.97} &  82.31 &   87.29 &  73.57 &  79.84  \\
{LSTM$^{ct_{a_{w+s}}}$+LSTM$^{pt_{a_{w+s}}}$} 	& 79.00	&	80.14	&	79.56	&	80.43	&	79.29	&	79.86 \\

\end{tabular}
\caption{Experimental results for $IAC_{{v2}_{st}}$ dataset using  prior and succeeding turns as context ({\bf bold} are best scores)}
\label{table:resultsv2st}
\end{table*}

\subsection{Error Analysis} \label{section:ea}

We conducted an error analysis of our models and identified the following types: 

\paragraph{{\bf Missing Background Knowledge}}

Sarcasm or verbal irony depends to a large extent upon the shared knowledge of the speaker and hearer (common ground) that is not explicitly part of the conversation context \citep{haverkate1990speech}. 
For instance, check the following context/sarcastic reply pair from the $IAC_{v2}$ corpus.
\begin{quote}
{\bf userA:} i'm not disguising one thing. I am always clear that my argument is equal marriage for same sex couples. No one i know on my side argues simply for ``equality in marriage''. \\
{\bf userB:} Right, expect when talking about the 14th amendment, The way you guys like to define ``equal protection'' would make it so any restriction is unequal.
\end{quote}

Here, {\bf userB} is sarcastic while discussing the 14th amendment (i.e., equal protection). On social media, users often argue about different topics, including controversial ones.\footnote{As stated earlier, $IAC$ includes a large set of conversations from 4forums.com, a website for political debates \citep{justo2014extracting,walker2012corpus}.} When engaged in conversations, speakers might assume that some background knowledge about those topics is understood by the hearers (e.g., historical events, constitution, politics). For example, posts from $Reddit$ are based on specific subreddits where users share similar interests (i.e., video games). We found, often, that even if the sarcastic posts are not political, they are based on specific shared knowledge (e.g., the performance of a soccer player in recent games). LSTM or SVM models are unable to identify the sarcastic intent when such contextual knowledge that is outside of the conversation context is used by the speaker. In future, however, we intend to build a model on specific subreddits (i.e., politics, sports) to investigate how much the domain-specific knowledge helps the classifiers.


\paragraph{{\bf Longer Sarcastic Reply}}
Although, the $IAC_{v2}$ and $Reddit$ corpora are annotated differently (using crowdsourcing vs. self-labeled, respectively), the labels are for the posts and not for specific sentences. Thus for longer posts, often the LSTM models perform poorly since the sarcastic cue is buried under the remaining non-sarcastic parts of the post. For instance, we observe that about 75\% of the false negative cases reported by the LSTM$^{c_{a_{s}}}$+LSTM$^{r_{a_{s}}}$ on the $IAC_{v2}$ data have 5 or more sentences in the sarcastic posts. 

\paragraph{{\bf Use of Profanity and Slang}} Sarcasm could be bitter, caustic, snarky or could have a mocking intent. \citet{orabycreating} asked the annotators to look for such characteristic while annotating the $IAC_{v2}$ posts for sarcasm. We observe that although the LSTM models are particularly efficient in identifying some inherent characteristics of sarcastic messages such as ``context incongruity'' (detailed in Section \ref{section:qualitative}), they often miss the sarcastic posts that contain slang and the use of profane words. In the future, we plan to utilize a lexicon to identify such posts, similar to \citep{burfoot2009automatic}.

\paragraph{{\bf Use of Numbers}} In some instances, sarcasm is related to situations that involve numbers, and the models are unable to identify such cases (i.e., userB: ``why not? My mother has been 39, for the last 39 years.'' in reply of userA: ``actually the earth is 150 years old. fact and its age never changes''). This type of sarcasm often occurs in social media both in discussion forums and on Twitter  \citep{joshi2015}.   

\paragraph{{\bf Use of Rhetorical Questions}}
We also found that sarcastic utterances that use rhetorical questions ($RQ$), especially in discussion forums (e.g., $IAC_{v2}$) are hard to identify. \citet{orabycreating} hypothesized that  sarcastic utterances of $RQ$ type are of the following structure: they contain questions in the middle of a post, that are followed by a statement. Since many discussion posts are long and might include multiple questions, question marks are not very strong indicators for $RQ$. 



\section{Qualitative Analysis} \label{section:qualitative}

\citet{wallace2014humans} showed that by providing additional conversation context, humans could identify sarcastic utterances which they were unable to do without the context. However, it will be useful to understand whether a specific \textit{part of the conversation context triggers} the sarcastic reply. To begin to address this issue, we conducted a qualitative study to understand whether (a) human annotators can identify parts of context  that trigger the sarcastic reply and (b) attention weights can signal similar information. For (a) we designed a crowdsourcing experiment (Crowdsourcing Experiment 1 in Section \ref{section:crowd}), and for (b) we looked at the attention weights of the LSTM networks (Section \ref{section:comparison}). 

In addition, discussion forum posts are usually long (several sentences), and we noticed in our error analysis that computational models have a harder time to correctly label them as sarcastic or not. The second issue we want to investigate is whether there is a particular sentence in the sarcastic post that expresses the speaker's sarcastic intent. To begin to address this issue, we conducted another qualitative study to understand whether (a) human annotators can identify a sentence in the sarcastic post that mainly expresses the speaker's sarcastic intent and (b) the sentence-level attention weights can signal similar information. For (a) we designed a crowdsourcing experiment (Crowdsourcing Experiment 2 in Section \ref{section:crowd}), and for (b) we looked at the attention weights of the LSTM networks (Section \ref{section:comparison}). 



For both studies, we compare the human annotators' selections with the attention weights to examine whether the attention weights of the LSTM networks are correlated to human annotations.

\subsection{Crowdsourcing Experiments} \label{section:crowd}

\begin{figure*}[t]
\centering
\subfloat[ \label{fig:t}]
 {\includegraphics[width=1.6in]{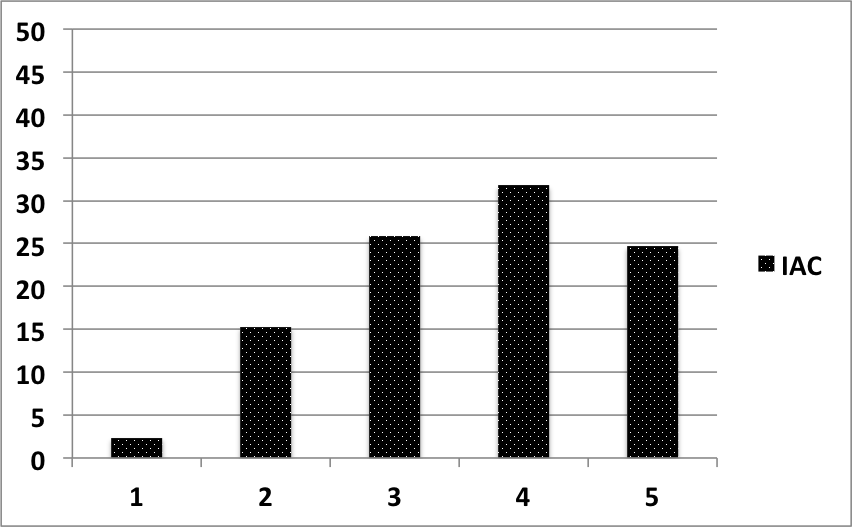}}
 \hspace{1cm}
\subfloat[ \label{fig:s}]
 {\includegraphics[width=1.6in]{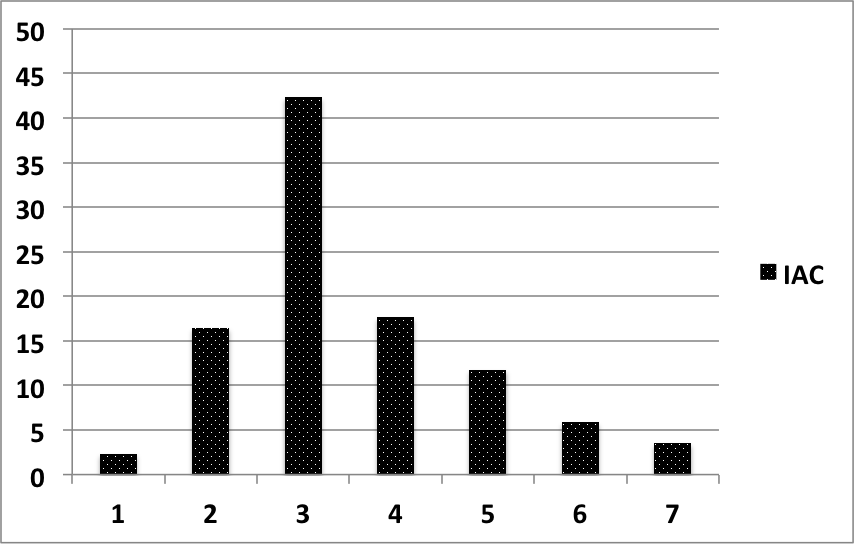}}
 \centering
 \caption{Crowdsourcing Experiment 1: (a) number of different trigger selections made by the five turkers (1 means all Turkers selected the exact same trigger(s)) and (b) distribution of the number of sentences chosen by the Turkers as triggers in a given post; both in \%}
 \label{figure:85ts}
\end{figure*}

\paragraph{{\bf Crowdsourcing Experiment 1}} We designed an Amazon Mechanical Turk task (for brevity, MTurk) as follows: given a pair of a sarcastic current turn ({\sc c\_turn}) and its prior turn ({\sc p\_turn}), we ask Turkers to identify one or more sentences in {\sc p\_turn} that they think triggered the sarcastic reply. Turkers could select one or more sentences from the conversation context {\sc p\_turn}, including the entire turn. We selected all sarcastic examples from the $IAC_{v2}$ test set where the prior turn contain between 3-7 sentences, since longer turns might be a more complex task for the Turkers. This selection resulted in 85 pairs. We provided several definitions of sarcasm to capture all characteristics. The first definition is inspired by the ``Standard Pragmatic Model'' \citep{grice1975syntax} that says verbal irony or sarcasm is a speech or form of writing which means the opposite of what it seems to say. In another definition, taken from \citet{orabycreating}, we mentioned that sarcasm often is used with the intention to mock or insult someone or to be funny. We provided a couple of examples of sarcasm from the $IAC_{v2}$ dataset to show how to successfully complete the task (See Appendix for the instructions given the the Turkers). Each HIT contains only one pair of {\sc c\_turn} and {\sc p\_turn} and five Turkers were allowed to attempt each HIT. Turkers with reasonable quality (i.e., more than 95\% of acceptance rate with experience of over 8,000 HITs) were selected and paid seven cents per task. Since Turkers were asked to select one or multiple sentences from the prior turn, standard inter-annotator agreement (IAA) metrics are not applicable. Instead, we look at two aspects to understand the user annotations. First, we look at the distribution of the triggers (i.e., sentences that trigger the sarcastic reply) selected by the five annotators (Figure \ref{figure:85ts}). It can be seen that in 3\% of instances all five annotators selected the exact same trigger(s), while in 58\% of instances 3 or 4 different selections were made per posts. Second, we looked at the distribution of the number of sentences in the {\sc p\_turn} that were selected as triggers by Turkers. We notice that in 43\% of time three sentences were selected. 

\begin{figure*}[t]
\centering
\subfloat[ \label{fig:t}]
 {\includegraphics[width=1.6in]{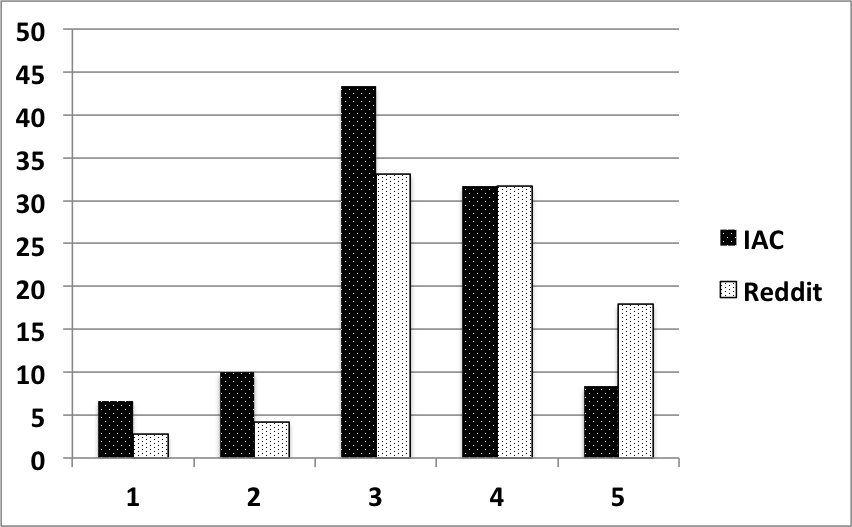}}
  \hspace{1cm}
\subfloat[ \label{fig:s}]
 {\includegraphics[width=1.6in]{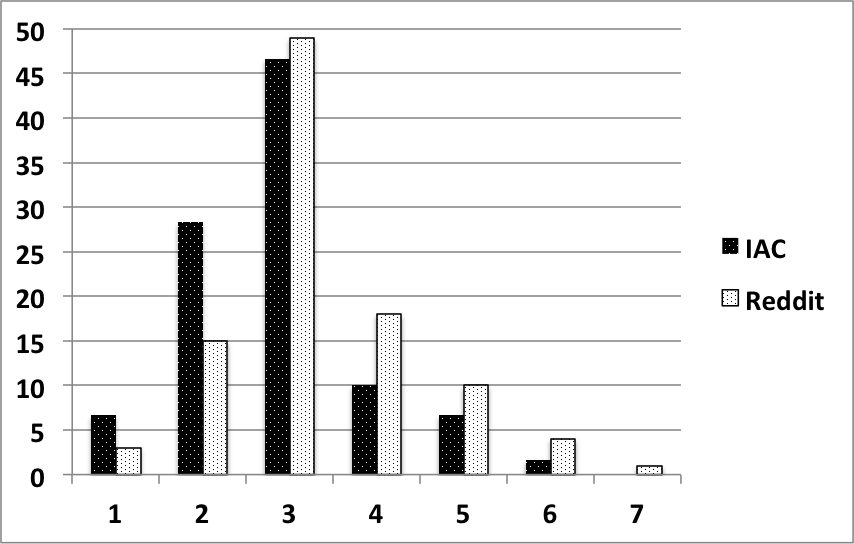}}
 \centering
 \caption{Crowdsourcing Experiment 2: (a) number of different trigger selections made by the five turkers (1 means all Turkers selected the exact same trigger(s)) and (b) distribution of the number of sentences chosen by the Turkers as triggers in a given post; both in \%}
 \label{figure:uniqts}
\end{figure*}

\paragraph{{\bf Crowdsourcing Experiment 2}} The second study is an extension of the first study. Given a pair of a sarcastic turn {\sc c\_turn} and its prior turn {\sc p\_turn}, we ask the Turkers to perform two subtasks. First, they were asked to identify ``only one'' sentence from {\sc c\_turn} that expresses the speaker's sarcastic intent. Next, based on the selected sarcastic sentence, they were asked to identify one or more sentences in {\sc p\_turn} that may trigger that sarcastic sentence (similar to the Crowdsourcing Experiment 1). 
We selected examples both from the $IAC_{v2}$ corpus (60 pairs) as well as the $Reddit$ corpus (100 pairs).  Each of the {\sc p\_turn} and {\sc c\_turn} contains three to seven sentences (note that the examples from the $IAC_{v2}$ corpus are a subset of the ones used in the previous experiment). We replicate the same design as the previous MTurk (i.e., we included definitions of sarcasm, provided examples of the task, use only one pair of {\sc c\_turn} and {\sc p\_turn} per HIT, required the same qualification for the Turkers, and paid the same payment of seven cents per HIT; See Appendix for the instructions given to Turkers).  Each HIT was done by five Turkers (a total of 160 HITs). To measure the IAA between the Turkers for the first subtask (i.e., identifying a particular sentence from {\sc c\_turn} that expresses the speaker's sarcastic intent) we used Krippendorf's $\alpha$ \citep{krippendorff2012content}. We measure IAA on nominal data, i.e., each sentence is treated as a separate category. Since the number of sentences (i.e., categories) can vary between three and seven, we report separate $\alpha$ scores based on the number of sentences. For {\sc c\_turn} that contains three, four, five or more than five sentences, the $\alpha$ scores are 0.66, 0.71, 0.65, 0.72, respectively. The $\alpha$ scores are modest and illustrate (a) identifying sarcastic sentence from a discussion forum post is a hard task and (b) it is plausible that the current turn ({\sc c\_turn}) contains multiple sarcastic sentences. For the second subtask, we carried a similar analysis as for experiment 1, and results are shown in Figure \ref{figure:uniqts} both for the $IAC_{v2}$ and $Reddit$ data.    

\subsection{Comparing Turkers' Answers with the Attention Weight of the LSTM models} \label{section:comparison}

In this section, we compare the Turkers' answers for both tasks with the sentence-level attention weights of the LSTM models. This analysis is an attempt to provide an interpretation of the attention mechanism of the LSTM models for this task.  

To identify what part of the prior turn triggers the sarcastic reply, we first measure the overlap of Turkers' choice with the sentence-level attention weights of the LSTM$^{ct_{a_{s}}}$+LSTM$^{pt_{a_{s}}}$ model. For Crowdsourcing Experiment 1, we used the models that are train/tested on the $IAC_{v2}$ corpus. We selected the sentence with the highest attention weight and matched it to the sentence selected by Turkers using majority voting. We found that 41\% of the times the sentence with the highest attention weight is also the one picked by Turkers. 
Figures \ref{figure:tongue} and \ref{figure:bible} show side by side the heat maps of the attention weights of LSTM models (LHS) and Turkers' choices when picking up sentences from the prior turn that they thought triggered the sarcastic reply (RHS). For Crowdsourcing Experiment 2, 51\% and 30\% of the times the sentence with the highest attention weight is also the one picked by Turker for $IAC_{v2}$ and $Reddit$, respectively.  

To identify what sentence of the sarcastic current turn expresses best the speaker's sarcastic intent, we again measure the overlap of Turkers' choice with the sentence-level attention weights of LSTM$^{ct_{a_{s}}}$+LSTM$^{pt_{a_{s}}}$ model (looking at the sentence-level attention weights from the current turn). We selected the sentence with the highest attention weight and matched it to the sentence selected by Turkers using majority voting. For $IAC_{v2}$, we found that 25\% of the times the sentence with the highest attention weight is also the one picked by the Turkers. For $Reddit$, 13\% of the times the sentence with the highest attention weight is also the one picked by the Turkers. The low agreement on $Reddit$ illustrates that many posts may contain multiple sarcastic sentences.

For both of these issues, the obvious question that we need to answer is why these sentences are selected by the models (and humans). In the next section, we conduct a qualitative analysis to try answering this question.

\subsection{Interpretation of the Turkers' Answers and the Attention Models}

We visualize and compare the sentence-level as well as the word-level attention weights of the LSTM models with the Turkers' annotations. 


\begin{figure}
\begin{center}
  \begin{minipage}[c]{0.40\linewidth}
\vspace{0pt}
    \centering
\begin{tabular}{ p{.5cm}|p{5cm} }\hline
S1 & Ok...  \\ 
S2 & I have to stop to take issue with something here,
 that I see all too often. \\ 
 S3 & And I've held my tongue on this as long as I can \\ 

\end{tabular}
\label{table:student}

\end{minipage}
 \hspace{1cm}
  \begin{minipage}[c]{0.40\linewidth}
\vspace{0pt}
    \includegraphics[width=0.9\textwidth]{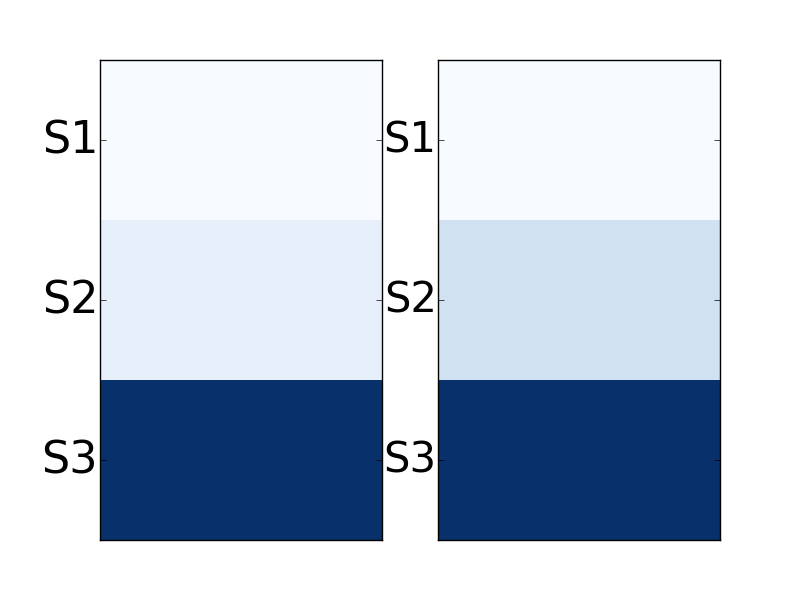}

  \end{minipage}%
\caption{Sentences in {\sc p\_turn}; heatmap of the \textit{attention weights} (LHS) and \textit{Turkers' selection} (RHS) of which of those sentences trigger the sarcastic {\sc c\_turn}=``Well, it's not as though you hold your tongue all that often when it serves in support of an anti-gay argument.''}
\label{figure:tongue}
\end{center}
\end{figure}

\begin{figure}
\begin{center}
  \begin{minipage}[c]{0.40\linewidth}
\vspace{0pt}
    \centering
\begin{tabular}{ p{.5cm}|p{6cm}}\hline
S1 & How do we rationally explain these creatures existence \dots for millions of years?  \\ 
S2 &  and if they were the imaginings of bronze age \dots we can now recognize from fossil evidence? \\ 
 S3 & and while your at it \dots 200 million years without becoming dust? \\ 

\end{tabular}
\label{table:student}

\end{minipage}
 \hspace{2cm}
  \begin{minipage}[c]{0.40\linewidth}
\vspace{0pt}
    \includegraphics[width=0.9\textwidth]{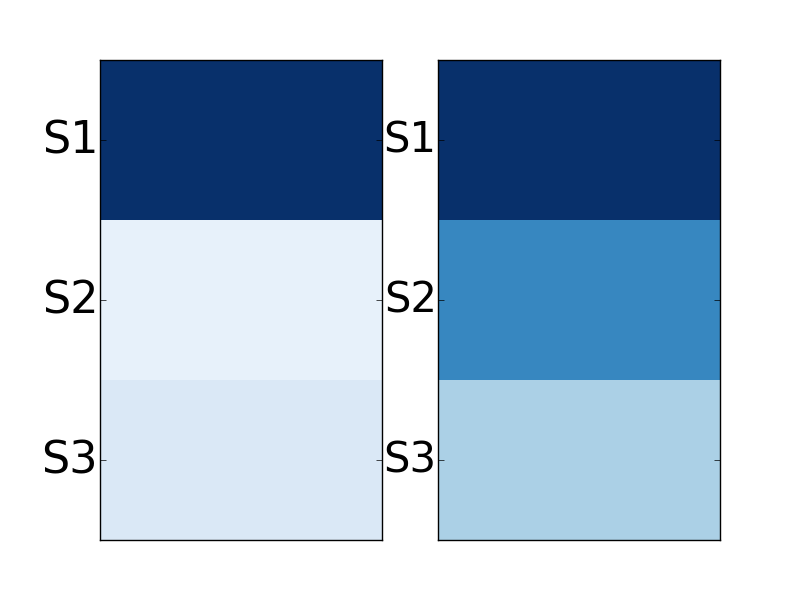}

  \end{minipage}%

\caption{Sentences in {\sc p\_turn} (userC in Table \ref{table:priorcurrent}); heatmap of the \textit{attention weights} (LHS) and \textit{Turkers' selection} (RHS) of which of those sentences trigger the sarcastic {\sc c\_turn} (userD in Table \ref{table:priorcurrent}). }
\label{figure:bible}
\end{center}
\end{figure}

\paragraph{Semantic coherence between prior turn and current turn}

Figure \ref{figure:tongue} shows a case where the prior turn contains three sentences, and the sentence-level attention weights are similar to the Turkers' choice of what sentence(s) triggered the sarcastic turn. Looking at this example it seems the model pays attention to output vectors that are \textit{semantically coherent} between {\sc p\_turn} and {\sc c\_turn}. The sarcastic {\sc c\_turn} of this example, contains a single sentence --
``Well, it's not as though you hold your tongue all that often when it serves in support of an anti-gay argument'', while the sentence from the prior turn {\sc p\_turn} that received the highest attention weight is S3 ``And I've held my tongue on this as long as I can''. 
                                                        
In Figure \ref{figure:bible}, the highest attention weight is given to the most informative sentence --``how do we rationally explain these creatures existence so recently in our human history if they were extinct for millions of years?''. Here, the sarcastic post {\sc c\_turn} (userD's post in Table \ref{table:priorcurrent}) mocks userC's prior post (``how about this explanation -- you're reading waaaaay too much into your precious bible''). For both the figures --- Figure \ref{figure:tongue} and Figure \ref{figure:bible}, the sentence from the prior turn {\sc p\_turn} that received the highest attention weight has also been selected by the majority of the Turkers. For Figure \ref{figure:tongue} the distribution of the attention weights and Turkers' selections are alike. Both examples are taken from the $IAC_{v2}$ corpus. 

\begin{figure}
  \begin{minipage}[c]{0.40\linewidth}
\vspace{0pt}
\centering
\begin{tabular}{p{1.1cm} |p{.5cm}|p{5.5cm}}\hline
& S1 & nothing will happen, \dots other private member motions.  \\ 
{\sc p\_turn} & S2 &  this whole thing is being made \dots are trying to push an agenda. \\ 
& S3 & feel free to let your \dots discussion before it is send to the trashcan. \\ 

\end{tabular}
\label{table:student}

\end{minipage}
 \hspace{2cm}
  \begin{minipage}[c]{0.40\linewidth}
\vspace{0pt}
    \centering
    \includegraphics[width=0.9\textwidth]{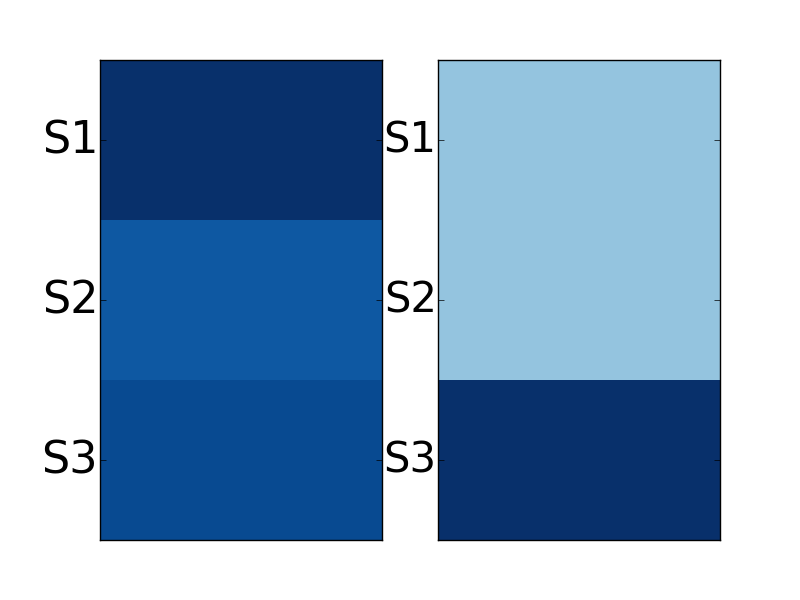}

  \end{minipage}%
\vspace{2ex}
  \begin{minipage}[c]{0.45\linewidth}
\vspace{0pt}
\centering
\begin{tabular}{p{1.1cm}  |p{.5cm}|p{8cm} }\hline
& S1 & the usual ``nothing to see here'' response.  \\ 
{\sc c\_turn} & S2 &  whew! \\ 
 & S3 & we can sleep at night and ignore this. \\ 

\end{tabular}
\label{table:student}

\end{minipage}
 \hspace{2cm}
  \begin{minipage}[c]{0.40\linewidth}
\vspace{0pt}
    \centering
    \includegraphics[width=0.9\textwidth]{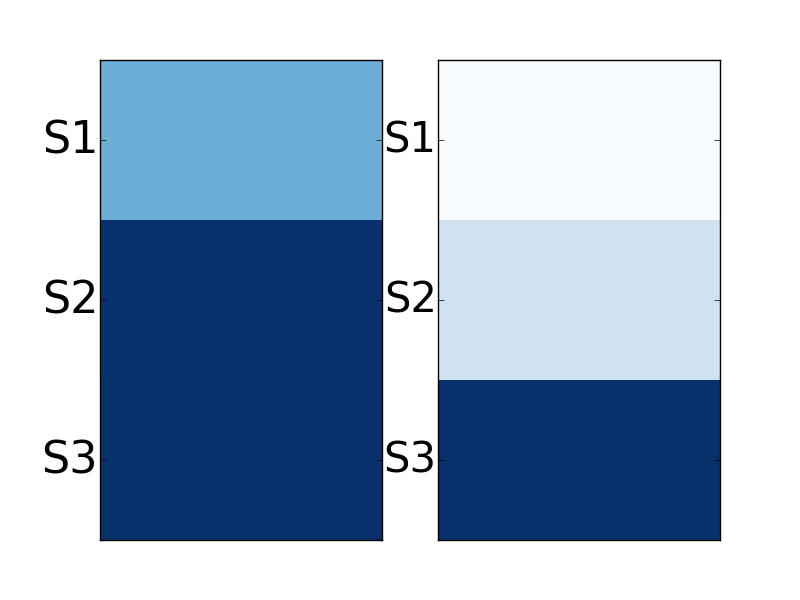}

  \end{minipage}%

\caption{Sentences from {\sc p\_turn} that trigger sarcasm (Top) and Sentences from {\sc c\_turn} that express sarcasm (Bottom). Tables show respectively the text from {\sc p\_turn} and {\sc c\_turn} (top and bottom) and Figure shows the heatmap of \textit{attention weights} (LHS) and \textit{Turkers' selection} (RHS)}
\label{figure:parliament}
\end{figure}

Figure \ref{figure:parliament} shows a pair of conversation context (i.e., prior turn) and the sarcastic turn (userE and userF's posts in Table \ref{table:priorcurrent}) together with their respective heatmaps that reflect the two subtasks performed in the second crowdsourcing experiment. The bottom part of the figure represents the sentences from the {\sc c\_turn} and the heatmaps that compares attention weights and the Turkers' selections for the first subtask: selecting the sentence from {\sc c\_turn} that best expresses the speaker's sarcastic intent. The top part of the figure shows the sentences from the {\sc p\_turn} as well as the heatmaps to show what sentence(s) are more likely to trigger the sarcastic reply. 
We make two observations: (a) Different Turkers selected different sentences from the {\sc c\_turn} as expressing sarcasm. 
The attention model has given the highest weight to the last sentence in {\sc c\_turn} similar to the Turkers's choice; (b) The attention weights seem to indicate semantic coherence between the sarcastic post (i.e, ``nothing to see here'' with the prior turn ``nothing will happen, this is going to die \dots''). 

\begin{figure}[h]
  \centering
  \subfloat[\label{fig:wordmap1}]
  {\includegraphics[width=1.7in]{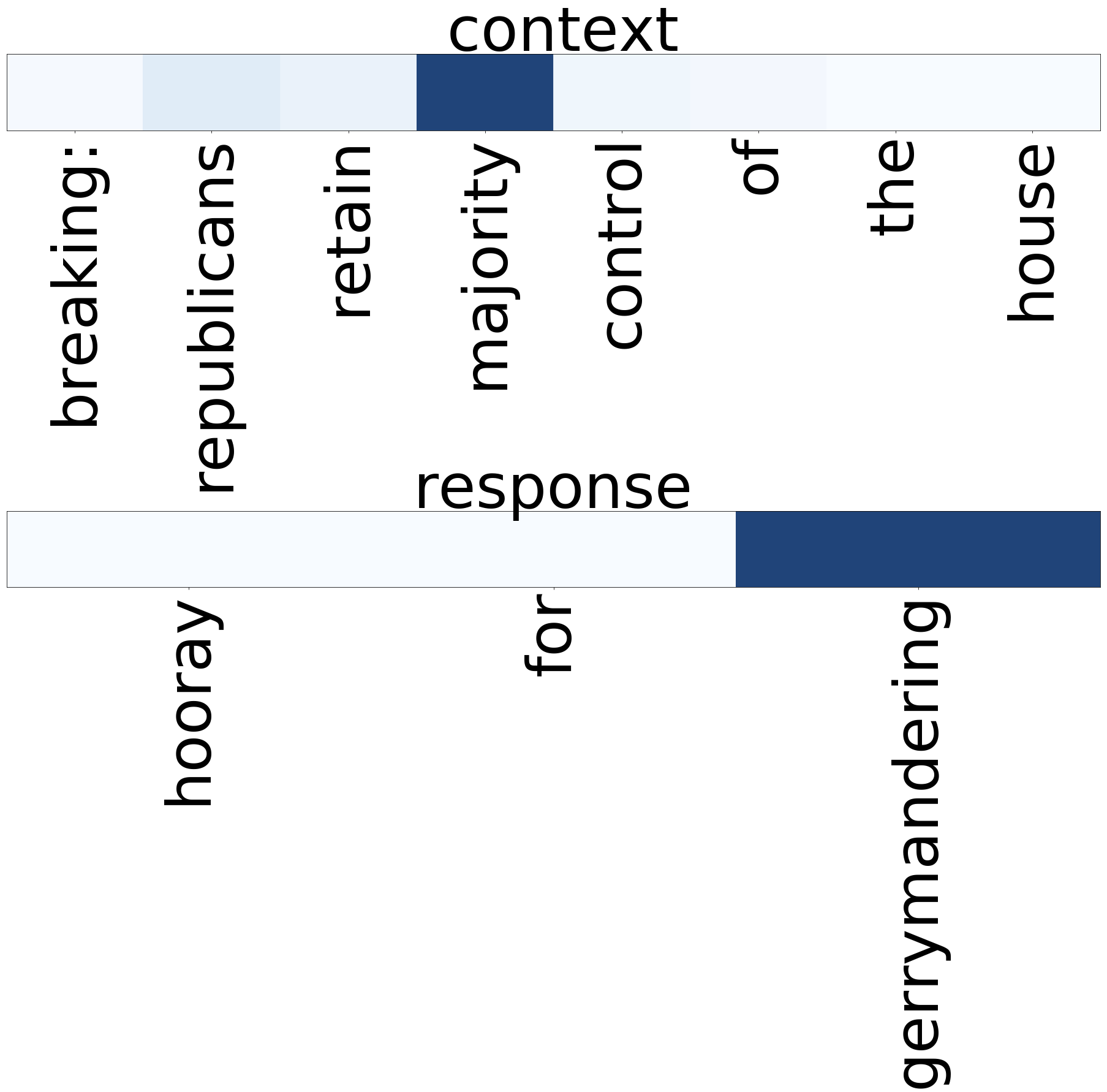}}
  \vfill
  \subfloat[\label{fig:wordmap2}
  ]
  {\includegraphics[width=1.7in]{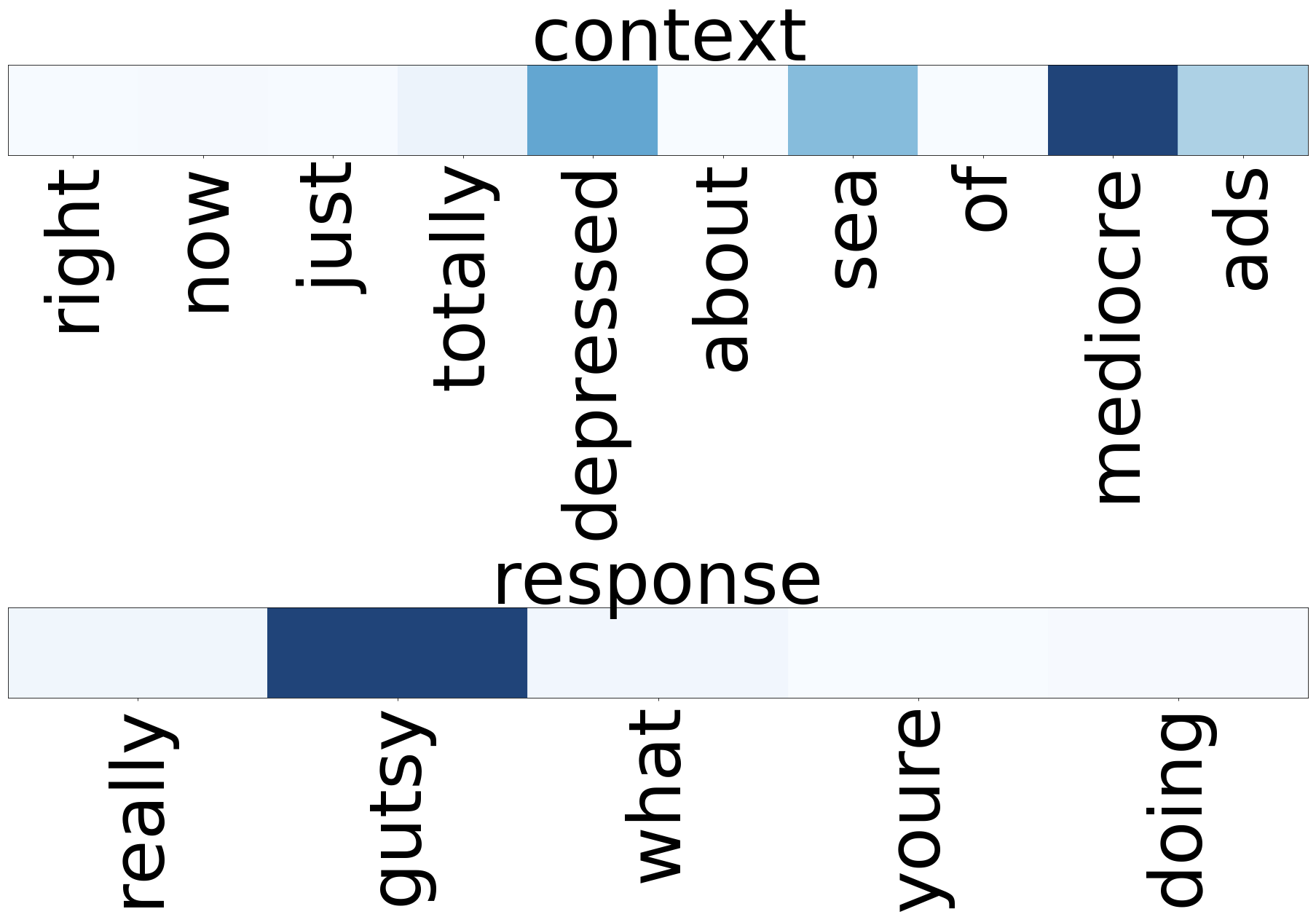}}
   \vfill
  \subfloat[\label{fig:wordmap3}
  ]
  {\includegraphics[width=1.7in]{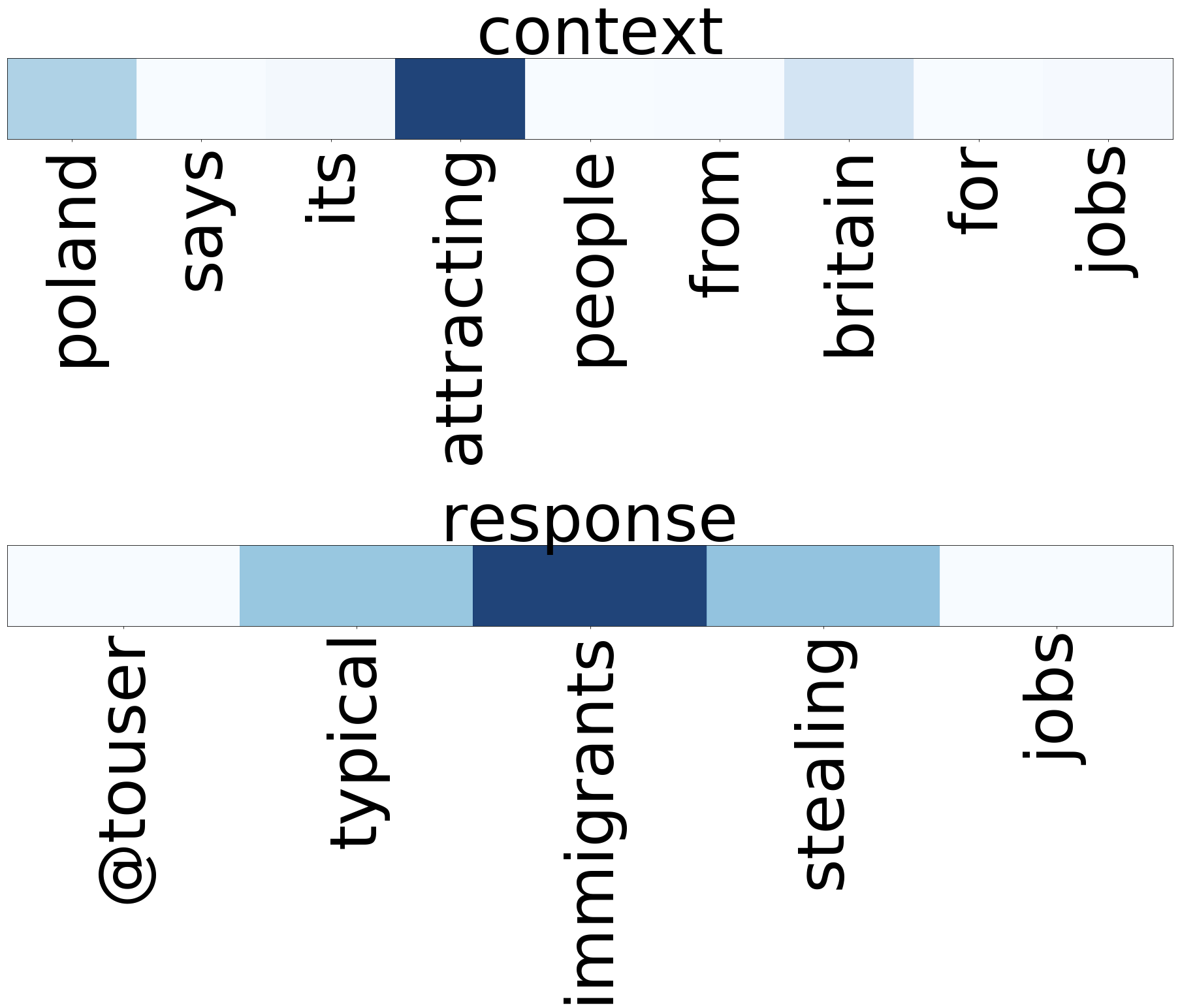}}
   \vfill
  \subfloat[\label{fig:wordmap4}
  ]
  {\includegraphics[width=1.7in]{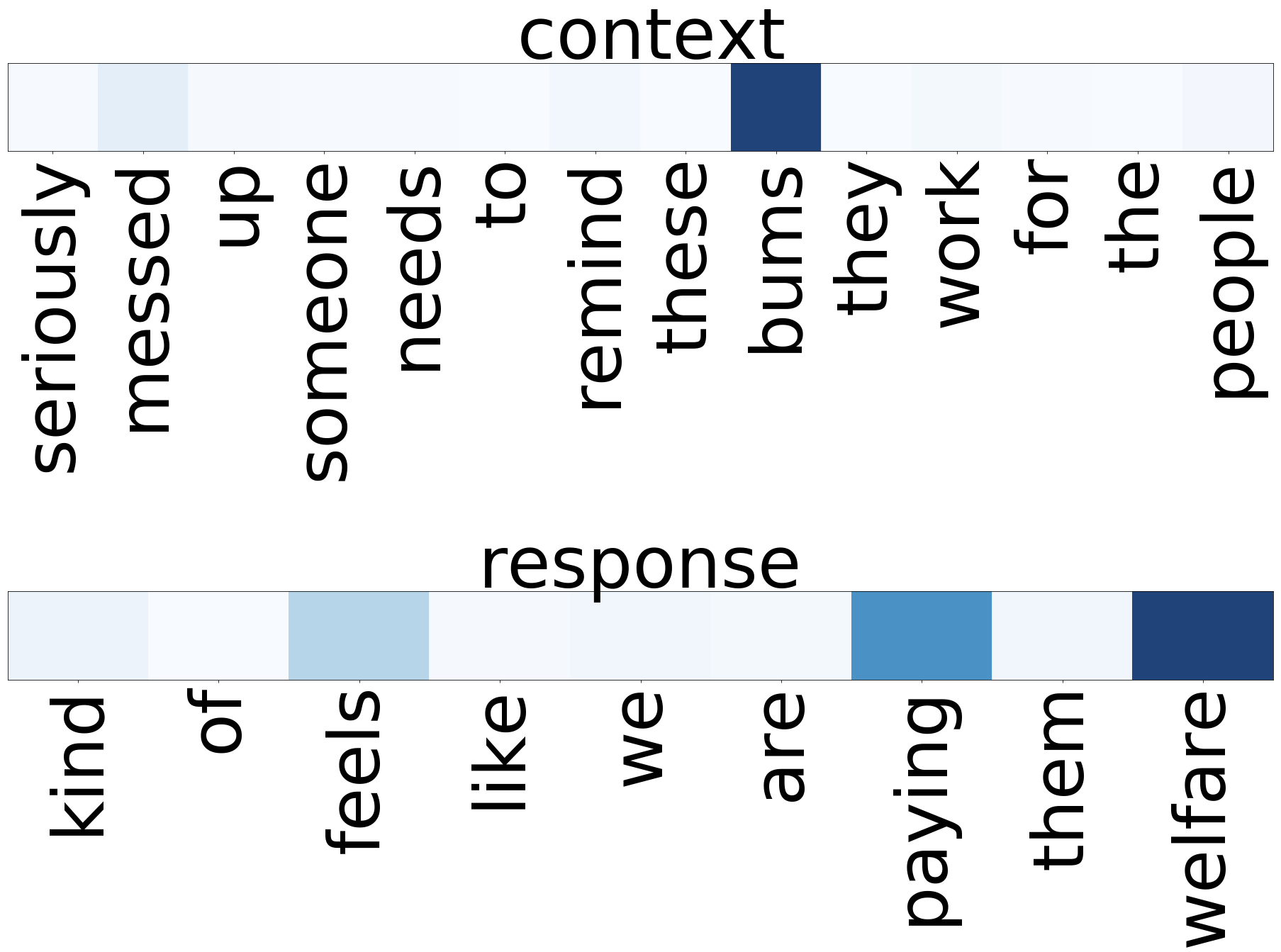}}
 \caption{Attention visualization of  incongruity between {\sc p\_turn}s and {\sc c\_turn}s on $Twitter$}
 \label{figure:twitterfigs}

\end{figure}

We also observe similar behavior in tweets (highest attention to words --\textit{majority} and \textit{gerrymandering} in Figure \ref{figure:twitterfigs}(d)). 


\begin{figure}
  \begin{minipage}[c]{0.40\linewidth}
\vspace{0pt}
\centering
\begin{tabular}{p{1cm} |p{.5cm}|p{5.5cm} }\hline
& S1 & technically speaking : this guy 
chose to fight in the ukraine.  \\ 
{\sc p\_turn} & S2 &  he died because of it. \\ 
& S3 & sure russia fuels the conflict,
 but he didnt have to go there. \\ 
& S4 & his choice, his consequences. \\ 
\end{tabular}
\label{table:student}

\end{minipage}
 \hspace{2cm}
  \begin{minipage}[c]{0.40\linewidth}
\vspace{0pt}
    \centering
    \includegraphics[width=0.9\textwidth]{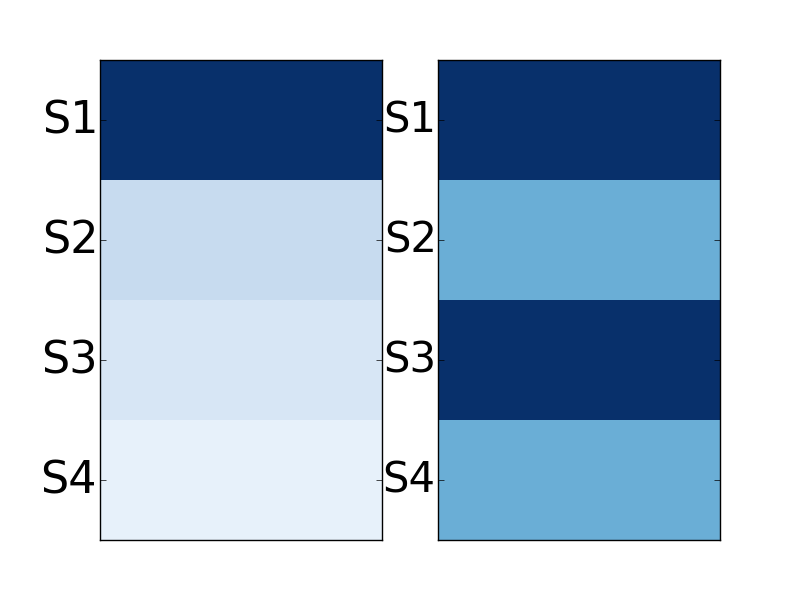}

  \end{minipage}%
\vspace{2ex}
  \begin{minipage}[c]{0.40\linewidth}
\vspace{0pt}
\centering
\begin{tabular}{p{1cm}  |p{.5cm}|p{5.5cm} }\hline

& S1 & sure thing.   \\ 
{\sc c\_turn} & S2 &  protecting your home from an looter? \\ 
 & S3 & nope, why would anyone do that?
 \\ 

\end{tabular}
\label{table:student}

\end{minipage}
 \hspace{2cm}
  \begin{minipage}[c]{0.40\linewidth}
\vspace{0pt}
    \centering
    \includegraphics[width=0.9\textwidth]{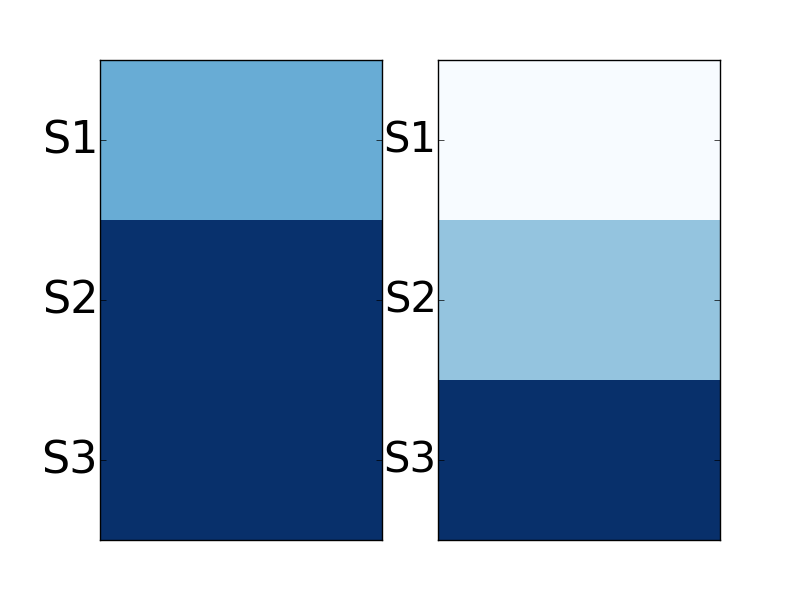}

  \end{minipage}%

\caption{Sentences from {\sc p\_turn} that trigger sarcasm (Top) and Sentences from {\sc c\_turn} that represents sarcasm (Bottom). Tables show respectively the text from {\sc p\_turn} and {\sc c\_turn} (top and bottom) and Figure shows \textit{attention weights} (LHS) and \textit{Turkers' selection} (RHS)}
\label{figure:ukraine}
\end{figure}

\paragraph{Incongruity between Conversation Context ({\sc p\_turn}) and Current Turn ({\sc c\_turn})}
Context incongruity is an inherent characteristic of irony and sarcasm and has been extensively studied in linguistics, philosophy, communication science \citep{grice1975syntax,attardo2000irony,burgers2012verbal} as well as recently in NLP \citep{riloff,joshi2015}. It is possible that the literal meaning of the current turn {\sc c\_turn} is incongruent with the conversation context ({\sc p\_turn}). We observe in discussion forums and Twitter that the attention-based models have frequently identified sentences and words from {\sc p\_turn} and {\sc c\_turn} that are semantically incongruous. For instance, in Figure \ref{figure:ukraine}, the attention model has given more weight to sentence S2 (``protecting your home from a looter?'') in the current turn, while from the {\sc p\_turn} the model assigned the highest weight to sentence S1 (`'`this guy chose to fight in the ukraine''). Here the model picked up the opposite sentiment from the {\sc p\_turn} and 
{\sc c\_turn},that is ``chose to fight'' and ``protecting home from looter''. Thus, the model seems to learn the incongruity between the prior turn {\sc p\_turn} and the current turn {\sc c\_turn} regarding the opposite sentiment. Also, the attention model selects (i.e., second highest weight) sentence S2 from the {\sc p\_turn} (``he died because of it'') which also shows the model captures opposite sentiment between the conversation context and the sarcastic post.  

However, from Figure \ref{figure:ukraine}, we noticed that some of the Turkers choose the third sentence S3 (``sure russia fuels the conflict, but he didnt have to go there'') in addition to sentence S1 from the context {\sc p\_turn}. Here, the Turkers utilize their background knowledge on global political conflicts (see Section \ref{section:ea}) to understand the context incongruity, fact missed by the attention model.

In the Twitter dataset, we observe that the attention models often have selected utterance(s) from the context which has opposite sentiment (Figure \ref{figure:twitterfigs}(a), Figure \ref{figure:twitterfigs}(b), and Figure \ref{figure:twitterfigs}(c)). Here, the word and sentence-level attention model have chosen the particular utterance from the context (i.e., the top heatmap for the context) and the words with high attention (e.g., ``mediocre'' vs. ``gutsy''). 
 Word-models seem to also work well when words in the prior turn and current turn are semantically incongruous but not related to sentiment (``bums'' and ``welfare'' in context: ``someone needs to remind these \textit{bums} they work for the people'' and reply: ``feels like we are paying them \textit{welfare}'' (Figure \ref{figure:twitterfigs}(d)).  


\paragraph{Attention weights and sarcasm markers}
Looking just at the attention weights in the replies, we notice the models are giving the highest weight to sentences that contain sarcasm markers, such as emoticons (e.g.,, ``:p'', ``:)'') and interjections (e.g., ``ah'', ``hmm''). We also observe that interjections such as ``whew'' with exclamation mark receive high attention weights (Figure \ref{figure:parliament}; see the attention heatmap for the current turn {\sc c\_turn}). Sarcasm markers such as the use of emoticons, uppercase spelling of words, or interjections, are explicit indicators of sarcasm that signal that an utterance is sarcastic \citep{attardo2000irony,burgers2012verbal,ghosh20181}. Use of such markers in social media (mainly on Twitter) is extensive. 

\paragraph{Reversal of valence}
The reversal of valence is an essential criterion of sarcastic messages that states that the intended meaning of the
sarcastic statement is opposite to its literal meaning \citep{burgers2010verbal}. One of the common ways of representing sarcasm is through a ``sarcastic praise'' (i.e., sarcasm with a
positive literal meaning as in ``Great game, Bob!'', when the game was poor) and
sarcastic blame (i.e., sarcasm with a negative literal meaning as in ``Horrible game, Bob!", when the game was great). \citet{ghoshguomuresan2015EMNLP} have studied the use of words that are used extensively in social media, particularly on Twitter to represent sarcastic praise and blame. For instance, words such as ``genius'', ``best'' are common in representing sarcastic praise since we need to alter their literal to intended meaning to identify the sarcasm. In our analysis, we observe, often the attention models have put the highest weights to such terms (i.e., ``greatest'', ``mature'' ,``perfect'') whose intended use in the sarcastic statement is opposite to its literal meaning.

\section{Conclusions and Future Directions}

This research makes a complementary contribution to existing work on modeling context for sarcasm/irony detection by looking at a particular type of context, \emph{conversation context}. We have modeled both the prior and succeeding turns when available as conversation context. Although Twitter is the de-facto platform for research on verbal irony or sarcasm we have thoroughly analyzed both Twitter and discussion forums datasets. 

We have addressed three questions: 

\begin{enumerate}
\item{\emph{Does modeling of conversation context help in sarcasm detection?} To answer this question, we show that only if we explicitly model the context and the current turn using a multiple-LSTM architecture, we obtain improved results as compared to just modeling the current turn. The multiple-LSTM architecture is designed to recognize a possible inherent incongruity between the current turn and the context, and thus is important to keep the C\_TURN and the context (P\_TURN and/or S\_TURN) separate as long as possible. This incongruity might become diffuse if the inputs are combined too soon, and we have shown that the multiple-LSTM architecture outperforms a single LSTM architecture that combines the current turn and the context. 
In particular,  LSTM networks with sentence-level attention achieved significant improvement when using prior turn as context for all the datasets (e.g., 6-11\% F1 for $IAC_{v2}$ and Twitter messages). Using the succeeding turn did not prove to be helpful for our datasets.} 
\item{\emph{Can humans and computational models determine what part of the conversation context (P\_TURN) triggered the sarcastic reply (C\_TURN)?} To answer this question, we conducted a qualitative study to understand whether (a) human annotators can identify parts of the context  that trigger the sarcastic reply and (b) the attention weights of the LSTM models can signal similar information. This study also constitutes an attempt to provide an interpretation of the attention mechanism of the LSTM models for our task. Our results show, in crowdsourcing experiment 1, for 41\% of the times the sentence with the highest attention weight is also the one picked by the Turkers.}
\item{\emph{Given a sarcastic post that contains multiple sentences is it feasible to identify a particular sentence that expresses the speaker's sarcastic intent?} To answer this question we conducted another qualitative study to understand whether (a) human annotators can identify a sentence in the sarcastic post that mainly expresses the speaker's sarcastic intent and (b) the sentence-level attention weights can signal similar information. This study again aims to provide an interpretation of the attention mechanism of the LSTM models. For this task, the agreement between the attention weights of the models and humans (using majority voting) is lower than for the previous task. However, the IAA between Turkers is also just moderate (alpha between 0.66-0.72), which shows that this is inherently a difficult task. It might be also the case that a post/turn is sarcastic in general and not a single sentence can be selected as being the only one sarcastic. }
\end{enumerate}

Our experiments showed that attention-based models can identify inherent characteristics of sarcasm (i.e., sarcasm markers and sarcasm factors such as context incongruity). We also conducted a thorough error analysis and indicated several types of errors: missing world knowledge, use of slang, use of rhetorical questions, and use of numbers. In future work, we plan to develop approaches to tackle these errors such as modeling rhetorical questions (similar to \citet{oraby2017you}), having a specialized approach to model sarcastic messages related to numbers, or using additional lexicon-based features to include slang. 

Although a few recent research papers have conducted experiments on discussion forum data we understand that there are many questions to address here. First, we show that self-labeled sarcastic turns (e.g., $Reddit$) are harder to identify compared to a corpus where turns are externally annotated (crowdsourced) (e.g., $IAC_{v2}$). We show that even if the training data in $Reddit$ is ten times larger that did not make much impact in our experiments. However, the $Reddit$ corpus consists of several subreddits so it might be interesting in the future to experiment with training data from a particular genre of subreddit (e.g., political forums). Second, during crowdsourcing, the Turkers are provided with the definition(s) of the phenomenon under study, which is not applicable in self-labeled corpora. It is unclear whether authors of sarcastic or ironic posts are using any specific definition of sarcasm or irony while labeling (and we see ironic posts labeled with the \#sarcasm hashtag). 

In future work we plan to study the impact of using a larger context such as the full thread in a discussion similar to \citet{zayats2017conversation}. This will also be useful in order to gain a broader understanding of the role of sarcasm in social media discussions (i.e., sarcasm as a persuasive strategy). We are also interested in utilizing external background knowledge to model sentiment about common situations (e.g., going to the doctor; being alone) or events (e.g., rainy weather) that users are often sarcastic about. 
\appendix
\section{Appendix}
\subsection{Mechanical Turk Instructions for Crowdsourcing Experiments 1}

\subsubsection{Identify what triggers a sarcastic reply}

Sarcasm is a speech or form of writing which means the opposite of what it seems to say. Sarcasm is usually intended to mock or insult someone or to be funny. People participating in social media platform, such as discussion forums are often sarcastic. In this experiment, a pair of posts ({previous post} and {sarcastic reply}) from an online discussion forum is presented to you. The {sarcastic reply} is a response to the {previous post}. However, given these posts may contain more than one sentence, often sarcasm in the sarcastic reply is triggered by only one or just a few of the sentences from the previous post.

Your task will be to identify the sentence/sentences from the previous post that triggers the sarcasm in the sarcastic reply. Consider the following pair of posts (sentence numbers are in "()").
\begin{itemize}
\item{\textbf{\textit{UserA: previous post}}}: (1) It's not just in case of an emergency. (2) It's for everyday life. (3) When I have to learn Spanish just to order a burger at the local Micky Dee's, that's a problem. (4) Should an English speaker learn to speak Spanish if they're going to Miami?
\item{\textbf{\textit{UserB: sarcastic reply}}}: When do you advocate breeding blond haired, blue eyed citizens to purify the US?
Here, the author of the sarcastic reply is sarcastic on previous post and the sarcasm is triggered by sentence 3 (``When I have to learn Spanish\dots'') and sentence 4 (``Should an English speaker\dots'') and not the other sentences in the post.
\end{itemize}

\textbf{DESCRIPTION OF THE TASK}

Given such a pair of online posts, your task is to identify the sentences from the previous post that trigger sarcasm in the sarcastic reply. You only need to select the sentence numbers from the previous post (do not retype the sentences). Here are some examples of how to perform the task.

\textbf{Example 1}
\begin{itemize}
\item{\textbf{\textit{UserA: previous post}}}: (1) see for yourselves. (2) The fact remains that in the caribbean, poverty and crime was near nil. (3) Everyone was self-sufficient and contented with the standard of life. (4) there were no huge social gaps.
\item{\textbf{\textit{UserB: sarcastic reply}}}: Are you kidding me?! You think that Caribbean countries are "content?!" Maybe you should wander off the beach sometime and see for yourself.
Answers - 2, 3.
\end{itemize}

\textbf{Example 2}
\begin{itemize}
\item{\textbf{\textit{UserA: previous post}}}: (1) Sure I can! (2) That is easy. (3) Bible has lasted thousands of years under the unending scrutiny of being judged by every historical discovery. (4) Never has it been shown to be fictional or false.
\item{\textbf{\textit{UserB: sarcastic reply}}}: Except for, ya know, like the whole Old Testament ;) False testament: archaeology refutes the Bible's claim to history.
\end{itemize}

Answers - 3, 4.

\subsection{Mechanical Turk Instructions for Crowdsourcing Experiments 2}

\subsubsection{Identify what triggers a sarcastic reply}
Sarcasm is a speech or form of writing which means the opposite of what it seems to say. Sarcasm is usually intended to mock or insult someone or to be funny. People participating in social media platforms, such as discussion forums are often sarcastic.

In this experiment, a pair of posts (previous post and sarcastic post) from an online discussion forum is presented to you. Suppose, the authors of the posts are respectively UserA and UserB. The sarcastic post from UserB is a response to the previous post from UserA. Your task is twofold. First, from UserB's sarcastic post you have to identify the particular "sentence" that presents sarcasm. Remember, you need to select only ONE sentence here. Next, given this sarcastic sentence look back at UserA's post. Often sarcasm in the sarcastic reply is triggered by only one or just a few of the sentences from the previous post. Your second task is to identify the sentence/sentences from the UserA's post that triggers the sarcasm in UserB's post.

Consider the following pair of posts (sentence numbers are in "()").

\begin{itemize}
\item{\textbf{\textit{UserA: previous post}}}: (1) see for yourselves. (2) The fact remains that in the caribbean, poverty and crime was near nil. (3) Everyone was self-sufficient and contented with the standard of life. (4) there were no huge social gaps.

\item{\textbf{\textit{UserB: sarcastic reply}}}: (1) Are you kidding me? (2) You think that Caribbean countries are ``content?'' (3) Maybe you should wander off the beach sometime and see for yourself.
Here, the sarcastic sentence in the sarcastic post of UserB is the 3rd sentence (``maybe you should wander off the beach\dots'')
\end{itemize}

At the same time, UserB is sarcastic on previous post from UserA and the sarcasm is triggered by sentence 2 ("Caribbean, poverty and crime was near nil \dots") and sentence 3 ("and everyone was self-sufficient \dots") and not the other sentences in the post.

\textbf{DESCRIPTION OF THE TASK}

Given such a pair of online posts, your task is twofold. First, you need to identify the sentence (i.e., only one sentence) from UserB's sarcastic reply that presents sarcasm. Next, from UserA's post select the sentences that trigger sarcasm in UserB's post. For both tasks you only need to select the sentence number (do not retype the sentences).

Here are some examples of how to perform the task.

Example 1
\begin{itemize}
\item{\textbf{\textit{UserA: previous post}}}: (1) Sure I can! (2) That is easy. (3) Bible has lasted thousands of years under the unending scrutiny of being judged by every historical discovery. (4) Never has it been shown to be fictional or false.
\item{\textbf{\textit{UserB: sarcastic reply}}}: (1) Except for, ya know, like the whole Old Testament ;) (2) False testament: archaeology refutes the Bible's claim to history.
\end{itemize}
Answers (Sentences triggering sarcasm) - 3, 4.\\
Answer (Sarcastic Sentence) - 1.

Example 2
\begin{itemize}
\item{\textbf{\textit{UserA: previous post}}}: (1) hasn't everyday since christ been latter days, thousands of days and he hasn't returned as promised. (2) in the bible his return was right around the corner ... how many years has it been. (3) when will you realize he isn't coming back for you!
\item{\textbf{\textit{UserB: sarcastic reply}}}: (1) how about when it dawns on you who he was when he came the first time? (2) lol (3) trade in your blinders for some spiritual light!
\end{itemize}
Answers (Sentences triggering sarcasm) - 1, 2, 3.\\
Answer (Sarcastic Sentence) - 3.

\section*{Acknowledgements}
This paper is based on work supported by the DARPA-DEFT program. The views expressed are those of the authors and do not reflect the official policy or position of the Department of Defense or the U.S. Government. The authors thank Christopher Hidey for the discussions and resources on LSTM and the anonymous reviewers for helpful comments. 

\starttwocolumn
\bibliography{acljournal2017conv}

\end{document}